\documentclass[twocolumn]{article}
\usepackage{lmodern} 
\usepackage[T1]{fontenc}
\usepackage{forest}
\usepackage{geometry}
\usepackage{setspace,lineno}

\usepackage{amsmath}
\usepackage{amsfonts}
\usepackage{amssymb}
\usepackage{float}
\usepackage{dblfloatfix} 
\usepackage{graphicx}
\usepackage{tabularx}
\usepackage{microtype}
\usepackage{cite}
\usepackage{url}
\usepackage{xcolor}
\usepackage{pifont}
\usepackage{authblk}
\newcolumntype{Y}{>{\centering\arraybackslash}X}
\newcommand{\cmark}{\textcolor{green!60!black}{\ding{51}}} 
\newcommand{\xmark}{\textcolor{red}{\ding{55}}}
\usepackage{booktabs}
\usepackage[colorlinks=true, linkcolor=blue, citecolor=blue, urlcolor=blue]{hyperref}

\title{From Language to Action: A Review of Large Language Models as Autonomous Agents and Tool Users}
\author{
Sadia Sultana Chowa\textsuperscript{1,*}, 
Riasad Alvi\textsuperscript{2}, 
Subhey Sadi Rahman\textsuperscript{2}, 
Md Abdur Rahman\textsuperscript{2}, 
Mohaimenul Azam Khan Raiaan\textsuperscript{2,*,†}, 
Md Rafiqul Islam\textsuperscript{3}, 
Mukhtar Hussain\textsuperscript{3}, 
Sami Azam\textsuperscript{3,*,†}\\
\small
\textsuperscript{1} Department of Computer Science and Engineering, Daffodil International University, Dhaka-1341, Bangladesh\\
\small
\textsuperscript{2}Department of Computer Science and Engineering, United International University, Dhaka 1212, Bangladesh\\
\small
\textsuperscript{3}Faculty of Science and Technology, Charles Darwin University, Casuarina, NT 0909, Australia\\
\small 
\textsuperscript{†}Equal Supervision. \\
\small 
\textsuperscript{*}Corresponding Author: mraiaan191228@bscse.uiu.ac.bd, sadia15-3052@diu.edu.bd, sami.azam@cdu.edu.au
}
\date{}

\begin{document}
	
\maketitle

\begin{abstract}

The pursuit of human-level artificial intelligence (AI) has significantly advanced the development of autonomous agents and Large Language Models (LLMs). LLMs are now widely utilized as decision-making agents for their ability to interpret instructions, manage sequential tasks, and adapt through feedback. This review examines recent developments in employing LLMs as autonomous agents and tool users and comprises seven research questions. We only used the papers published between 2023 and 2025 in conferences of the A* and A-ranked and Q1 journals. A structured analysis of the LLM agents' architectural design principles, dividing their applications into single-agent and multi-agent systems, and strategies for integrating external tools is presented. In addition, the cognitive mechanisms of LLMs, including reasoning, planning, and memory, and the impact of prompting methods and fine-tuning procedures on agent performance are also investigated. Furthermore, we have evaluated current benchmarks and assessment protocols and provided an analysis of 68 publicly available datasets to assess the performance of LLM-based agents in various tasks. In conducting this review, we have identified critical findings on verifiable reasoning of LLMs, the capacity for self-improvement, and the personalization of LLM-based agents. Finally, we have discussed ten future research directions to overcome these gaps.

\end{abstract}

\vspace{0.5em}


\section{Introduction}
Large language models (LLMs) have become central in artificial intelligence (AI) research due to their strong human-like ability to understand, generate, and reason in natural language \cite{kumar2024llms, wang2025llmwatermarking}. LLMs were used primarily as tools to serve as text generators or understanding modules within a larger application. However, further techniques such as few-shot prompting \cite{NEURIPS2020_1457c0d6}, chain-of-thought (CoT) prompting \cite{10.5555/3600270.3602070}, and self-ask prompting \cite{press-etal-2023-measuring} demonstrated how the potential of LLMs could be improved through smart prompting and input pattern design. Beyond conventional natural language processing (NLP) tasks, LLMs are now serving as autonomous agents and intelligent tools. They are embedded into increasingly complex workflows where they perform planning, decision making, and tool interaction in various real-world applications, including research assistance \cite{schmidgall2025agentlaboratoryusingllm}, software development \cite{qian2024chatdevcommunicativeagentssoftware}, drug discoveries \cite{gottweis2025aicoscientist}, multi-robot systems \cite{chen2025emosembodimentawareheterogeneousmultirobot}, clinical support \cite{CHEN2025151}, game simulation \cite{mao2024alympicsllmagentsmeet} and scientific simulations \cite{park2023generativeagentsinteractivesimulacra}. 

LLMs as agents can observe their environment, make decisions, and take actions. Within this paradigm, single-agent LLM systems have demonstrated promising performance in decision-making tasks. Single-agent systems such as Reflexion \cite{shinn2023reflexion}, Toolformer \cite{schick2023toolformer}, and ReAct \cite{yao2023react} showed how models can operate in decision loops that involve planning, memory, and tool use. However, they often struggle in dynamic environments that require simultaneous context tracking, external memory integration, and adaptive tool usage \cite{10970024, gao2025agent}. To address these limitations, the concept of multi-agent LLM systems has gained increasing attention. In such systems, multiple LLMs interact as specialized agents, each with distinct roles or goals, collaborating to solve more complex tasks than a single agent can manage. Through structured communication, reflective reasoning, and explicit role assignments in simulated settings, multi-agent LLMs exhibit capabilities such as consensus building, uncertainty-aware planning, and autonomous tool interaction \cite{talebirad2023multiagentcollaborationharnessingpower, zhang-etal-2024-exploring, han2025llmmultiagentsystemschallenges}. Examples such as MetaGPT \cite{hong2023metagpt}, CAMEL \cite{li2023camel}, AgentBoard \cite{chang2024agentboard}, AutoAct \cite{qiao2024autoactautomaticagentlearning}, and ProAgent \cite{zhang2024proagent} showcase how cooperative agents execute role-specific instructions and coordinate plans, while Generative Agents \cite{park2023generativeagentsinteractivesimulacra} simulate human-like behaviors in interactive environments.

\begin{figure*}[ht!]
  \centering
  \includegraphics[scale=0.2]{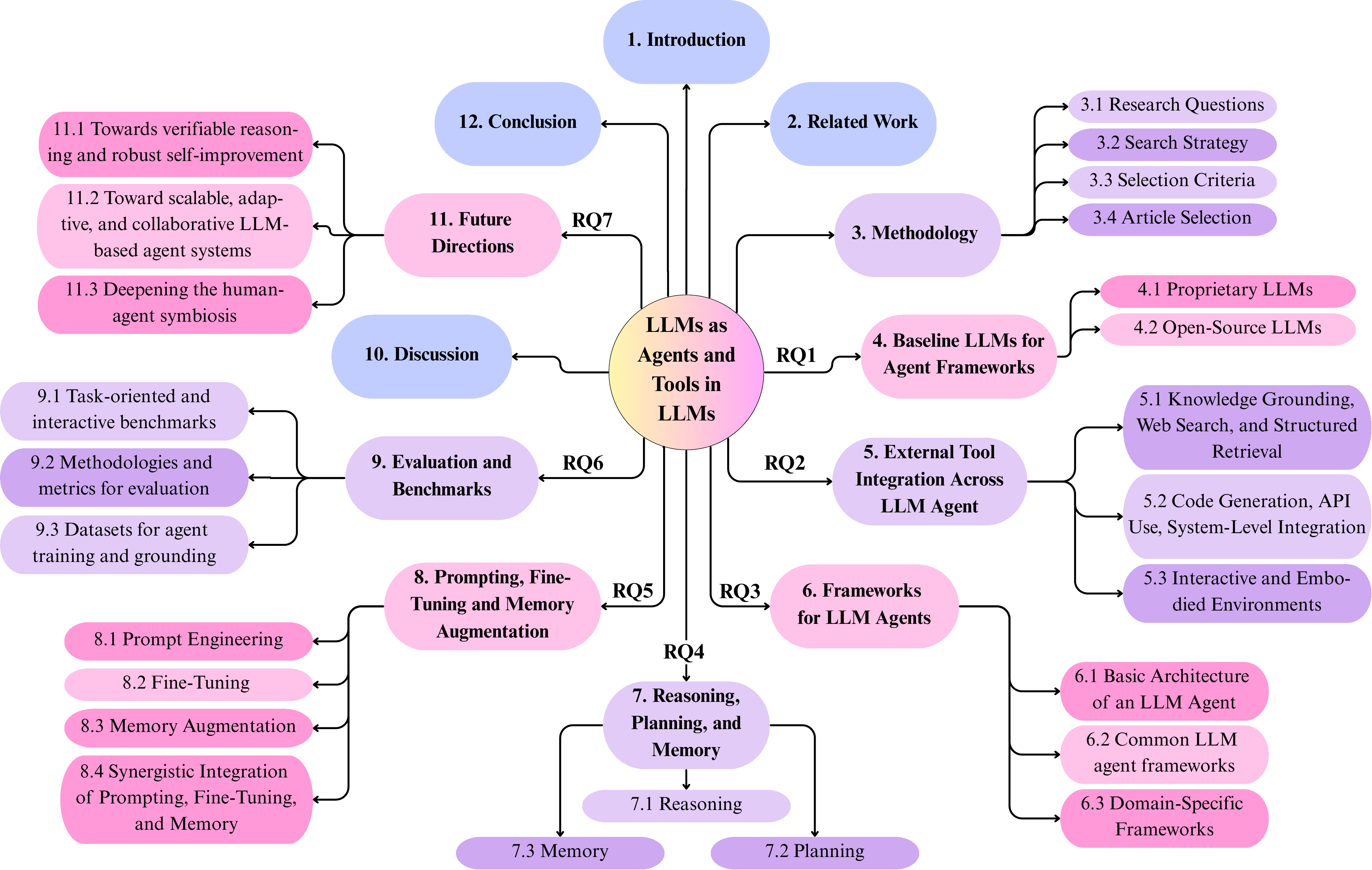}
  \caption{An overview of the taxonomy used in this review.}
  \label{fig:taxonomy}
\end{figure*}

\begin{table*}[ht!]
\centering
\scriptsize
\caption{Comparative analysis of existing survey papers on LLM agents and tool use based on key research questions}
\label{tab:literature}
\begin{tabularx}{\linewidth}{l *{7}{Y}}
\toprule
\textbf{Papers} & \textbf{RQ1} & \textbf{RQ2} & \textbf{RQ3} & \textbf{RQ4} & \textbf{RQ5} & \textbf{RQ6} & \textbf{RQ7} \\
\midrule
Ferrag et al. \cite{ferrag2025llmreasoningautonomousai} & \xmark & \cmark & \cmark & \cmark & \xmark & \cmark & \cmark \\
Li et al. \cite{li2025review} & \xmark & \cmark & \cmark & \cmark & \xmark & \xmark & \cmark \\
Xu et al. \cite{xu2025llmagents} & \cmark & \cmark & \xmark & \cmark & \xmark & \cmark & \cmark \\
Xi et al. \cite{xi2025llmagentsurvey} & \cmark & \cmark & \cmark & \cmark & \xmark & \xmark & \cmark \\
Wang et al. \cite{wang2024llmautonomous} & \xmark & \cmark & \cmark & \cmark & \xmark & \cmark & \cmark \\
Guo et al. \cite{guo2024largelanguagemodelbased} & \xmark & \cmark & \cmark & \xmark & \xmark & \cmark & \cmark \\
Cheng et al. \cite{cheng2024exploringlargelanguagemodel} & \xmark & \cmark & \cmark & \cmark & \xmark & \cmark & \cmark \\
\textbf{Ours} & \cmark & \cmark & \cmark & \cmark & \cmark & \cmark & \cmark \\
\bottomrule
\end{tabularx}
\end{table*}

Moreover, LLMs as agents and tools now demonstrate massive potential in AI, and the demand to understand their evolving roles has intensified. Therefore, a systematic review of its recent advancement, a discussion of the remaining gaps, and a research direction for future advancements are essential to advance the field. With this focus, this survey provides a comprehensive and structured overview of current capabilities and system designs. We investigate the architectural foundations that enable agent-like behavior in LLMs, analyze how they interact with external tools, discuss the key limitations of current approaches, and highlight the remaining open challenges. Through this survey, our objective is to map the landscape of this emerging field and to offer a solid foundation for future research and development.

\begin{table}[ht!]
  \scriptsize
  \caption{Taxonomy of LLM-based agentic systems}
  \centering
  \label{tab:tab_taxonomy}
    \begin{tabularx}{\columnwidth}{p{0.37\textwidth}>{\fontsize{5.7}{7}\selectfont\raggedright\arraybackslash}p{0.15\textwidth}}
    \toprule
    \textbf{Category} & \textbf{Ref.} \\
    \midrule
    \textbf{1. Core Methodologies and Agent Architectures} & \\

    \quad \textit{1.1 Multi-Agent Systems \& Collaboration Frameworks} & \\
    \quad\quad 1.1.1 General Collaborative Architectures & \cite{ni2024mechagents,chen2023agentverse,klein2025fleet,jin2024decoagent} \\
    \quad\quad 1.1.2 Domain-Specific Collaborative Architectures & \cite{lu2024triageagent,kim2024mdagents} \\
    \quad\quad 1.1.3 Hierarchical \& Role-Based Collaboration & \cite{bai2025collaboration,zhang2024chain} \\

    \quad \textit{1.2 Training \& Learning Paradigms} & \\
    \quad\quad 1.2.1 Reinforcement \& Self-Evolutionary Learning & \cite{ma2024coevolving} \\
    \quad\quad 1.2.2 Offline \& Self-Improvement Methods & \cite{xi2025llmagentsurvey,zhou2024star} \\
    \quad\quad 1.2.3 Modular \& Unified Training Architectures & \cite{yin-etal-2024-agentlumos,xu2024lemur} \\
    \quad\quad 1.2.4 Bootstrapping from LLM Knowledge & \cite{zhu2024bootstrapping,jorgensen2024large_gpagents} \\

    \quad \textit{1.3 Advanced Reasoning \& Planning Mechanisms} & \\
    \quad\quad 1.3.1 Structured \& Logical Reasoning & \cite{abdaljalil-etal-2025-theorem} \\
    \quad\quad 1.3.2 Planning with World Knowledge & \cite{qiao2024agent} \\
    \quad\quad 1.3.3 Contrastive Reasoning for Optimization & \cite{wu2024avatar} \\

    \addlinespace
    \textbf{2. Agent Capabilities and Enhancements} & \\

    \quad \textit{2.1 Tool \& API Integration} & \\
    \quad\quad 2.1.1 Frameworks for Tool Mastery & \cite{qin2024toolllm,yuan-etal-2025-easytool} \\
    \quad\quad 2.1.2 Tool-Augmented Reasoning in Specific Domains & \cite{RCAgent_10.1145/3627673.3680016}, \cite{goodell2025large,m2024augmenting} \\

    \quad \textit{2.2 Embodied Agents \& Physical/Virtual Interaction} & \\
    \quad\quad 2.2.1 Vision-Language Navigation (VLN) \& Grounding & \cite{schumann2024velma,zheng2024steveeye,hong2024cogagent}\\
    \quad\quad 2.2.2 Robotics \& Multi-Robot Task Planning & \cite{zhang2024building,10802322_SMART-LLM} \\
    \quad\quad 2.2.3 Unified Multimodal Interaction & \cite{wang2024omnijarvis} \\

    \quad \textit{2.3 Communication Mechanisms} & \\
    \quad\quad 2.3.1 Novel Communication Modalities & \cite{zhang2024generative} \\
    \quad\quad 2.3.2 Facilitating Agent Dialogue \& Negotiation & \cite{zhang2024large,li2023camel} \\

    \quad \textit{2.4 Personalization \& User Understanding} & \\
    \quad\quad 2.4.1 Implicit Intent Recognition & \cite{qian-etal-2024-tellmemore} \\
    \quad\quad 2.4.2 Personalized Agent Behavior & \cite{singh2024personal} \\

    \addlinespace
    \textbf{3. Domain-Specific Applications} & \\

    \quad \textit{3.1 Science \& Engineering} & \\
    \quad\quad 3.1.1 Scientific Research \& Discovery & \cite{ni2024mechagents,zhang2023large,ghafarollahi2024protagents,boiko2023autonomous} \\
    \quad\quad 3.1.2 Industrial \& Infrastructure Management & \cite{jin2024large,xia2024generation,xu2024large} \\
    \quad\quad 3.1.3 Electronic Design Automation (EDA) & \cite{wu2024chateda} \\

    \quad \textit{3.2 Healthcare \& Biomedicine} & \\
    \quad\quad 3.2.1 Clinical Decision Support \& Diagnosis & \cite{lu2024triageagent,kim2024mdagents,ColaCare_10.1145/3696410.3714877,ferber2025development,chen2025enhancing} \\
    \quad\quad 3.2.2 Biomedical Data Analysis \& Research & \cite{huang2024protchat,liu2025drbioright} \\
    \quad\quad 3.2.3 Patient \& Provider Communication & \cite{yang2024talk2care,Understanding_10.1145/3544548.3581503,abbasian2025conversational,barra2025prompt} \\
    \quad\quad 3.2.4 Medical Data Generation \& Calculation & \cite{goodell2025large} \\

    \quad \textit{3.3 Software, Code \& IT Operations} & \\
    \quad\quad 3.3.1 Code Generation \& Refinement & \cite{bai2025collaboration} \\
    \quad\quad 3.3.2 Test Case Generation & \cite{jorgensen2024large_gpagents} \\
    \quad\quad 3.3.3 Cloud Root Cause Analysis (RCA) & \cite{RCAgent_10.1145/3627673.3680016} \\

    \quad \textit{3.4 Economics, Finance \& Urban Planning} & \\
    \quad\quad 3.4.1 Macroeconomic \& Market Simulation & \cite{li-etal-2024-econagent,CompeteAI_10.5555/3692070.3694596} \\
    \quad\quad 3.4.2 Urban Knowledge Graph Construction & \cite{ning2024urbankgent} \\

    \quad \textit{3.5 Interactive Systems \& User Interfaces} & \\
    \quad\quad 3.5.1 Conversational Recommendation & \cite{Recommender_AI_10.1145/3731446} \\
    \quad\quad 3.5.2 Natural User Interfaces (Gesture \& Voice) & \cite{yang2024talk2care,zeng2024gesturegpt} \\

    \addlinespace
    \textbf{4. Evaluation, Safety, and Alignment} & \\

    \quad \textit{4.1 Benchmarking \& Evaluation Frameworks} & \\
    \quad\quad 4.1.1 General Agent Evaluation Platforms & \cite{wijk2025rebench,chang2024agentboard}\\
    \quad\quad 4.1.2 Domain-Specific Benchmarks & \cite{zhang-etal-2024-timearena,cheng2024sociodojo} \\

    \quad \textit{4.2 Safety, Security \& Robustness} & \\
    \quad\quad 4.2.1 Security Threats \& Backdoor Attacks & \cite{yang2024watch} \\
    \quad\quad 4.2.2 Privacy Preservation & \cite{zhang2024privacyasst} \\
    \quad\quad 4.2.3 Resilience to Faulty Agents & \cite{huang2025ontheresilience} \\
    \quad\quad 4.2.4 Safeguarding \& Guardrail Mechanisms & \cite{xiang2025guardagent} \\

    \quad \textit{4.3 Alignment \& Behavior Control} & \\
    \quad\quad 4.3.1 Value Alignment \& Social Norms & \cite{wang-etal-2024-sotopia,Selfalignment_10.5555/3692070.3693667} \\
    \quad\quad 4.3.2 Eliciting \& Mitigating Undesirable Behaviors & \cite{li2025eliciting} \\
    \quad\quad 4.3.3 Automated Guideline Generation & \cite{fu2024autoguide} \\

    \quad \textit{4.4 Understanding Agent Limitations \& Weaknesses} & \\
    \quad\quad 4.4.1 Probing for Failure Modes & \cite{george-etal-2024-probing,Understanding_10.1145/3544548.3581503} \\

    \addlinespace
    \textbf{5. Human-Agent Interaction and Social Dynamics} & \\

    \quad \textit{5.1 Human-in-the-Loop \& Collaboration} & \\
    \quad\quad 5.1.1 Synergistic Task Solving & \cite{feng-etal-2024-large} \\
    \quad\quad 5.1.2 Integrating with Symbolic AI for Explainability & \cite{frering2025integrating} \\

    \quad \textit{5.2 Simulating Human \& Social Phenomena} & \\
    \quad\quad 5.2.1 Modeling Social Cognition \& Prosocial Behavior & \cite{li-etal-2023-theoryofmind,zhang-etal-2024-exploring,liu2025exploring} \\
    \quad\quad 5.2.2 Simulating User Behavior \& Economic Dynamics & \cite{jin2024decoagent,UserBehavior_10.1145/3708985,behari2024decision} \\
    \midrule    
  \end{tabularx}
\vspace{-15pt}
\end{table}

Our key contributions are summarized as follows.
\begin{itemize}
    \item We conduct a comprehensive review of recent advancements in using LLMs as agents and tool users, with an explicit taxonomy that describes their architectures, frameworks, and interaction paradigms.
    \item We examine LLM reasoning, planning, and memory capabilities, and analyze how prompting, fine-tuning, and memory enhancement enhance agentic performance.
    \item We critically review current evaluation methods and benchmarks for LLM agents and tool users.
    \item We identify fundamental challenges, including alignment, reliability, and generalization, and outline promising research avenues to advance the robustness and intelligence of LLM agents.
\end{itemize}

The rest of the review is organized as follows. Section \ref{2_related_works} presents related works, identifying gaps in existing surveys and situating our contribution. Section \ref{3_methodology} outlines the methodology, including research questions, selection criteria, and search strategies. Section \ref{4_baseline} explores the baseline LLMs used in agentic LLM systems. Section \ref{5_external_tool} focuses on tool integration in LLM workflows. Section \ref{6_framework4agents} reviews the frameworks for constructing single-agent and multi-agent systems. Section \ref{sec:plan_reason} investigates the reasoning, planning, and memory capabilities of LLM agents. Section \ref{8_impacts} discusses prompting, fine-tuning, and memory augmentation techniques that enhance agentic behavior. Section \ref{9_eval} evaluates current benchmarks and assessment methodologies. Section \ref{10_discussion} provides a discussion. Section \ref{11_future_dir} outlines potential future research directions, and Section \ref{12_concl} concludes the review with a summary of the insights and contributions. Figure \ref{fig:taxonomy} illustrates the overall structure of the paper.

\section{Related works}
\label{2_related_works}
This section comprehensively analyzes the existing survey literature on LLMs as agents and tool users. The emergence of LLMs as autonomous agents and tool users has sparked interest in AI research. For example, Ferrag et al. \cite{ferrag2025llmreasoningautonomousai} provided a basic taxonomy of agents based on LLM, describing reasoning, planning, and tool use capabilities. Their survey cataloged over 60 benchmarks and systematically reviewed frameworks for agent behavior. Li et al. \cite{li2025review} also offered an analysis of three agentic paradigms: tool use, retrieval-based planning, and feedback-driven learning. They categorized LLM agent roles, discussed the limitations of task-agnostic frameworks, and proposed directions for composable and generalizable agent development. Similarly, Xu et al. \cite{xu2025llmagents} focused on tool-augmented LLMs and outlined strategies for integrating external functionalities, including prompting, multimodal interaction, and agent coordination. 

On the other hand, Xi et al. \cite{xi2025llmagentsurvey} conceptualized LLM agents within a modular architecture 'brain, perception, and action' that includes reasoning, planning, and tool interaction. Wang et al. \cite{wang2024llmautonomous} organized a unified agent framework that integrates core modules such as reasoning, memory, planning, and action control. Their survey reviews capability acquisition strategies and discusses how LLM agents engage with external tools. Guo et al. \cite{guo2024largelanguagemodelbased} examined LLM-based multi-agent systems, classified popular architectures and communication strategies, tools integration, and evaluated agent interaction through benchmarks. Cheng et al. \cite{cheng2024exploringlargelanguagemodel} analyzed the reasoning, planning, memory, and tool use mechanisms in single- and multi-agent environments. They explored architectural choices, prompting and fine-tuning techniques, and benchmark methodologies, while identifying limitations in adaptivity, robustness, and evaluation fidelity.

Although the existing surveys underscore significant progress in understanding LLM-based agents, particularly in tool use and architecture, they exhibit notable gaps in addressing the choice of baseline LLMs in multi-agent frameworks, the impact of prompting and fine-tuning, and a unified treatment of reasoning, memory, and evaluation. Addressing these limitations, our review systematically covers all key dimensions. 

Table~\ref{tab:literature} presents a comparative analysis of seven prominent surveys against our review content, structured around the research questions (RQs): baseline LLMs used (RQ1), integration of external tools (RQ2), frameworks for building LLM agents (RQ3), reasoning, planning, and memory capabilities (RQ4), prompting and fine-tuning strategies (RQ5), evaluation and benchmarks (RQ6) and concerns and limitations (RQ7).

Existing studies often emphasize the discussion of integrating external tools across LLM agent workflows (i.e., RQ2); however, foundational dimensions, such as baseline LLM usage and the impact of prompting, fine-tuning, and memory enhancement, receive comparatively limited attention. In contrast, addressing all seven key areas, we propose our own taxonomy of agentic systems based on LLM, presented in Table \ref{tab:tab_taxonomy}. This taxonomy extends existing frameworks by organizing the field into core methodologies, agent capabilities, domain-specific applications, evaluation and safety aspects, and human-agent interaction. Our holistic approach distinguishes us as the most comprehensive to date, addressing the foundational and emergent dimensions of agents and tools based on LLM, offering a unified perspective on their architectures, capabilities, and future directions.

\section{Methodology} 
\label{3_methodology}
This study uses a clear and organized methodology to explore the evolving field of LLM agents. The analysis is guided by targeted RQs that aim to clarify the basic structures, capabilities, and environments of these agents. The literature selection process included a wide range of studies that focused on foundational structures, new methods, and practical implementation approaches. The selected works were organized to allow for a detailed look at new trends, system designs, and mechanisms that allow agent-like behaviors in LLMs.

\subsection{Research questions (RQs)}
The main goal of this review is to synthesize the current state of LLM-based agents by examining their basic principles and real-world applications. To achieve this, the following research questions were created:
\begin{figure*}[!ht]
    \centering
    \includegraphics[scale=0.51]{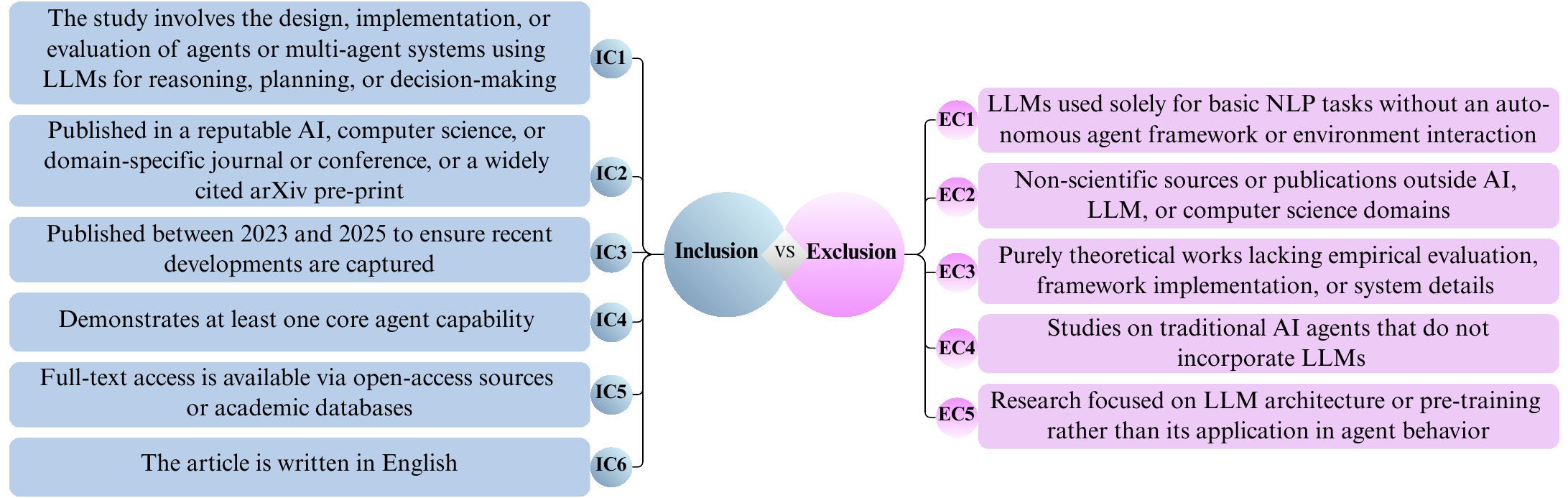}
    \caption{Inclusion and exclusion criteria for article selection}
    \label{fig:InclusionExclusion}
\end{figure*}
\begin{figure*}[ht!]
    \centering
    \includegraphics[scale=0.52]{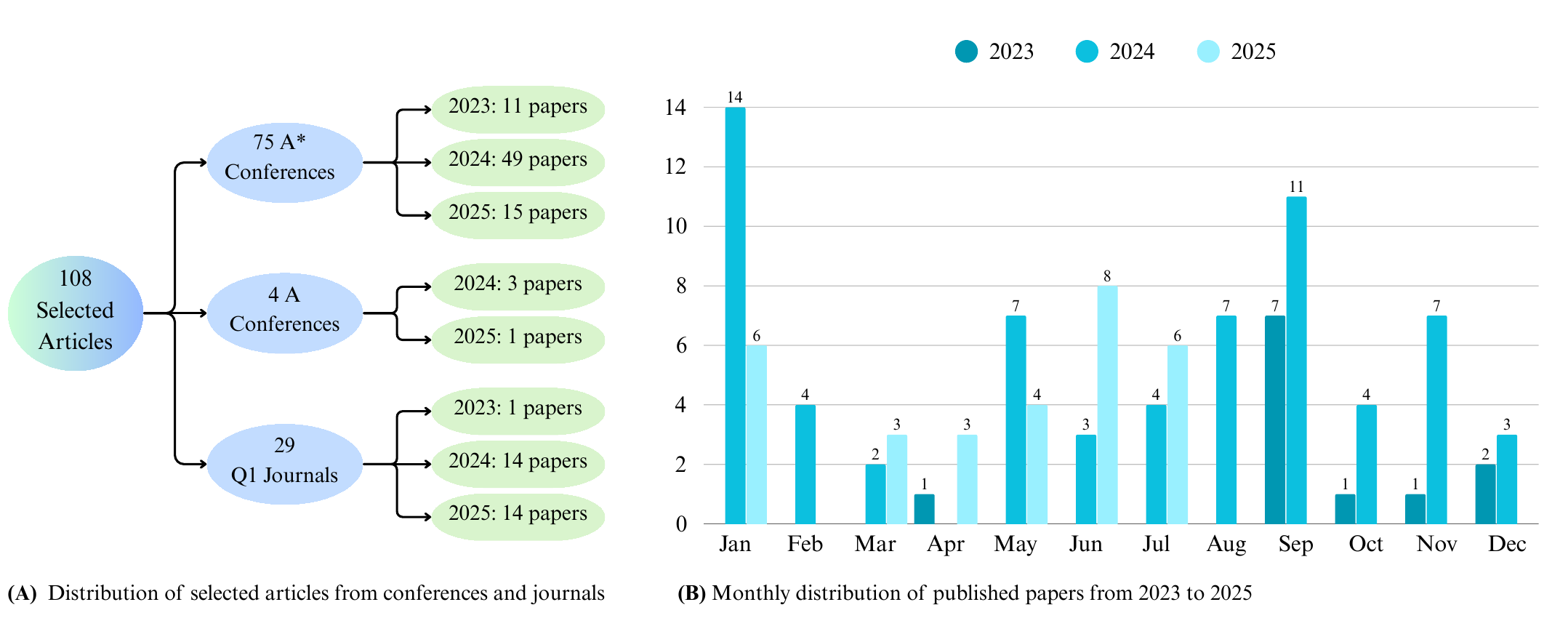}
    \caption{(A) Flow diagram illustrating the distribution of selected articles across conferences and journals. (B) Bar chart showing the monthly publication trends from 2023 to 2025}
    \label{fig:selected_articles}
\end{figure*}

\noindent \textbf{RQ1:} What core architectures and training mechanisms enable LLMs to exhibit agent-like behavior?

\noindent \textbf{RQ2:} How do LLMs interface with external tools, and what frameworks or paradigms govern this interaction?

\noindent \textbf{RQ3:} What are the key frameworks and systems for building single- or multi-agent ecosystems using LLMs?

\noindent \textbf{RQ4:} In what ways can LLM agents demonstrate reasoning, planning, memory, and self-reflection, and how do they compare with classical agents?

\noindent \textbf{RQ5:} How do prompting techniques, fine-tuning strategies, and memory augmentation impact the use and autonomy of tools in LLM agents?

\noindent \textbf{RQ6:} How is the performance of LLM agents evaluated, and what are the key benchmarks, metrics, and methodologies for measuring agent intelligence?

\noindent \textbf{RQ7:} What are the main challenges, limitations, and ethical concerns associated with the development and deployment of LLM-based agents?

These RQs are designed to offer a comprehensive multi-sided analysis of the field. RQ1 looks at the basic structures and training methods that help LLMs evolve from passive language models to active agents. RQ2 investigates how LLM agents interact with their environments, focusing on methods to integrate tools that guide their actions. RQ3 reviews the software frameworks and systems used in the creation and use of LLM agents, providing information on practical applications. RQ4 examines cognitive functions similar to those of LLM agents, such as reasoning, planning, and memory. RQ5 explores how optimization methods, including prompting, fine-tuning, and memory enhancement, affect agent independence and effectiveness. RQ6 addresses the critical area of evaluation, surveying the benchmarks and metrics used to validate agent capabilities and measure their effectiveness against established standards or human performance. In conclusion, RQ7 provides a critical perspective by investigating the inherent challenges, risks, and ethical considerations, such as reliability, security, and potential misuse, that accompany the rise of autonomous agents.

\subsection{Search strategies}
\textbf{Search sources.} The literature search focused on peer-reviewed publications from high-impact journals and conference proceedings in AI, machine learning, and natural language processing. The main sources included NeurIPS, ICML, ICLR, ACL, EMNLP, AAAI, EAAI, CVPR, ICCV, ACM, Nature Machine Intelligence, NPJ Digital Medicine, ACM Transactions, IEEE Transactions, and AI journals. These venues were selected for their reliable coverage of recent advances in LLMs and agent-based systems. The search looked for publications from 2023 to the present to capture the latest developments in this fast-moving field.

\textbf{Search terms.} A focused set of search terms was created to target the overlap of LLM and autonomous agents. These included: “Large Language Model agents,” “LLM-based agents,” “multi-agent LLM systems,” “tool-augmented LLMs,” “LLM planning and reasoning,” “LLM self-reflection,” “autonomous LLM agents,” “communicative LLM agents,” “LLM agent frameworks,” “embodied LLM agents,” and “LLM tool integration.” The terms were gradually refined on the basis of keywords found in the literature and emerging trends in the literature.

\subsection{Selection criteria}

To compile the final list of articles for this review, we set up a strict checklist of inclusion and exclusion criteria. The key inclusion criteria and exclusion criteria are shown in Figure \ref{fig:InclusionExclusion}.

\begin{table}[ht!]
\centering
\caption{Geographical distribution of publications in agentic LLM research}
\begin{scriptsize}
\renewcommand{\arraystretch}{1.15}
\label{tab:geo}
\begin{tabular}{>{\raggedright\arraybackslash}p{0.12\textwidth} >{\centering\arraybackslash}p{0.08\textwidth} >{\raggedright\arraybackslash}p{0.1\textwidth} >{\centering\arraybackslash}p{0.08\textwidth}}
\hline
\textbf{Country} & \textbf{Number of papers} & \textbf{Country} & \textbf{Number of papers} \\
\hline
China & 46 & Austria & 1 \\
USA & 42 & Korea & 1 \\
Germany & 5 & Denmark & 1 \\
UK & 3 & Ireland & 1 \\
Canada & 3 & Sweden & 1 \\
Singapore & 2 & Italy & 1 \\
Switzerland & 1 & & \\
\hline
\end{tabular}
\end{scriptsize}
\end{table}

\begin{table*}[ht!]
\centering
\caption{Overview of proprietary LLMs for LLM agent research}
\begin{scriptsize}
\begin{tabular}{p{4.1cm}p{4.7cm}p{5cm}p{3.5cm}}
    \midrule
    \textbf{Provider} & \textbf{Model Variants} & \textbf{Primary Applications} & \textbf{Key Differences} \\
    \midrule
    OpenAI \cite{li2025llm,jin2024large,zhang-etal-2024-benchmarking-data,Selfalignment_10.5555/3692070.3693667,ning2024urbankgent,chang2024agentboard,ruan2024identifying,bai2025collaboration,li-etal-2023-theoryofmind,lu2024triageagent,kim2024mdagents,bo2024reflective,ghafarollahi2024protagents,ni2024mechagents,chen2023agentverse,ma2024coevolving,Offline_10.5555/3692070.3694566,zhang-etal-2024-exploring,li-etal-2024-econagent,feng-etal-2024-large,george-etal-2024-probing,yuan-etal-2025-easytool,zhu2024bootstrapping,cheng2024sociodojo,huang2025ontheresilience,Understanding_the_Weakness_10.1145/3637528.3671650,jorgensen2024large_gpagents,cuadra2024digital,yang2024talk2care,zeng2024gesturegpt,Recommender_AI_10.1145/3731446,arumugam2025toward,huang2023benchmarking,zhang2024large,CompeteAI_10.5555/3692070.3694596,jin2024decoagent,zhang2024generative,frering2025integrating,yang2024llm,song2023llm,fu2024autoguide,zhang2024privacyasst,feldt2023towards,wei2024editable,xue2025comfybench,huang2024protchat,boiko2023autonomous,chen2025enhancing,m2024augmenting,ferber2025development,Describe_10.5555/3666122.3667602,zhang2023large,yang2024embodied,wijk2025rebench} & GPT-4, GPT-3.5, GPT-2, Text-DaVinci-003, Code-DaVinci-002, o1-preview, o1-mini & 
    Core reasoning, planning, benchmark reference, multi-agent collaboration, embodied agents & multimodal, tool-use/agents, advanced reasoning \\
    Anthropic \cite{zhang2024chain,chang2024agentboard,10802322_SMART-LLM,zhang2024privacyasst,wu2024avatar,liu2025exploring,wijk2025rebench,cai2025llm,li2025eliciting,zhang-etal-2024-exploring,ruan2024identifying,huang2023benchmarking} & Claude 1, 2, 3 Opus, 3.5, 3.7 Sonnet, 3 Haiku & 
    Benchmarking, time series learning, hallucination detection, R\&D comparison, behavior elicitation & ultra-long context, safety-first alignment, strong instruction following \\
    Google DeepMind \cite{zhang2024privacyasst, barra2025prompt,zhang-etal-2024-timearena,zhang-etal-2024-benchmarking-data,kim2024mdagents,behari2024decision,george-etal-2024-probing,liu2025exploring,zhang2024chain,muthusamy-etal-2023-towards} & Gemini 1.0 Pro, Gemini 1.5 Pro, Gemini Pro, PaLM 2, FLAN-T5 & 
    Agent development, distraction reasoning, comparative evaluation, enterprise agent foundation & native multimodal, million-token context, Google-ecosystem integration \\
    Others \cite{liu2025exploring,lampinen2023passive, Understanding_10.1145/3544548.3581503} & Mistral-medium-2312, Chinchilla-LM (70B), HyperCLOVA (82B) & 
    Benchmark tasks, collaborative agents, large-scale agent frameworks & inference-efficient small models, data-efficient scaling \\
    \midrule
\end{tabular}
\end{scriptsize}
\label{tab:proprietary_llm_agents_table}
\end{table*}
\subsection{Article selection}

We systematically reviewed agentic LLM systems and the role of integrated tools within such systems. Articles were selected from A* and A-ranked conferences and leading Q1 journals. Figure~\ref{fig:selected_articles} presents two complementary views of the publication distribution: (A) the distribution of selected articles between conferences and journals, and (B) the monthly distribution of published articles from 2023 to 2025. The review considered studies published between 2023 and 2025, with the majority appearing in 2024 and 2025. A total of 108 articles were included based on predefined inclusion (IC) and exclusion (EC) criteria. Among those, 75 were published in A* conferences, highlighting our concentration of high-impact research. Furthermore, Table~\ref{tab:geo} summarizes the geographical distribution of the publications, indicating that China and the United States account for the majority of contributions.  

\subsection{Thematic scoring for article selection}

A total of 108 papers were included in this review. To ensure a systematic and objective selection, each paper was evaluated using a structured thematic scoring framework, which quantified coverage across eight predefined dimensions:

\begin{itemize}
    \item Focus on LLM Agentic Behavior
    \item Training Mechanisms for Autonomy
    \item Interaction with External Tools or Environments
    \item Frameworks for Single/Multi-Agent Systems
    \item Reasoning, Planning, Memory, or Self-Reflection
    \item Prompting, Fine-Tuning, or Memory Augmentation
    \item Evaluation Metrics and Benchmarks for Agentic Performance
    \item Challenges, Limitations, and Ethical Considerations
\end{itemize}

Each dimension was scored on a scale from 0, 1, or 2 per co-author, where (0) means "no", (1) means "partially", and (2) means "yes". The score for each question is a cumulative score of four co-authors, with a maximum score of 8 for each question (i.e., if co-author one gives 2, co-author two gives 1, co-author three gives 0, and co-author four gives 2, then the thematic score of this question is 5). Then the total score for each paper was computed as the sum across all eight dimensions, with a maximum possible score of 64. Papers with higher total scores were considered more comprehensive, while all included papers met a minimum inclusion threshold (total score $\geq$ 48).


The scoring was conducted independently by four co-authors with expertise in LLMs and autonomous systems. Any disagreements in scoring were resolved through discussion until consensus was reached, ensuring high reliability and reproducibility. This structured approach allows for transparent, objective synthesis of the literature, highlighting methodological trends, key contributions, and gaps in the study of LLMs as autonomous agents.

The complete thematic scoring for all 108 papers is provided as supplementary material in Table S1.\footnote{\url{https://github.com/mak-raiaan/LLMAgentsReview}}


\section{Baseline LLMs for agent frameworks}
\label{4_baseline}

The architecture and functionality of an LLM agent are fundamentally related to its underlying language model \cite{cheng2024exploringlargelanguagemodel, CHEN2025151, su-etal-2024-language}. The selected foundational model ultimately determines the agent's attainable performance levels, associated costs, and flexibility for adaptation \cite{su-etal-2024-language, guo2024largelanguagemodelbased, zhang-etal-2024-benchmarking-data}. Our findings from the reviewed literature reveal a distinct pattern of model adoption, characterized by the widespread use of proprietary models alongside an increasing use of competitive open source models \cite{li2025eliciting, jin2024large, ma2024coevolving, zhang-etal-2024-timearena, li2025llm}. This section provides a structured overview of the LLMs employed across the reviewed papers, drawing attention to key trends in model selection, ranging from direct deployment to extensive fine-tuning.
\vspace{-10pt}

\subsection{Proprietary LLMs for agentic applications}

Several contemporary studies on agents are heavily based on state-of-the-art proprietary models from leading AI research organizations \cite{bai2025collaboration, cai2025llm}. These models are frequently chosen for their advanced reasoning, strong instruction-following abilities, and robust integration with external tools, positioning them as a reliable benchmark for evaluating novel agentic architectures and methods \cite{Offline_10.5555/3692070.3694566, wang-etal-2024-sotopia, Recommender_AI_10.1145/3731446, ruan2024identifying, zheng2024steveeye, zhang2024large}. Table \ref{tab:proprietary_llm_agents_table} provides a comprehensive summary of the proprietary LLMs examined in this review.

Among the available language models, the Generative Pre-trained Transformer (GPT) family, most notably GPT-4 and its variants, remains the most widely adopted foundation for agent implementations \cite{li2025llm}. These models often serve as the core decision-making component or serve as reference baselines to assess the relative performance of novel techniques. For example, GPT models such as GPT-4 and GPT-3.5 provide a consistent point of comparison when evaluating the effectiveness of other open source LLMs \cite{li2025llm, jin2024large, zhang-etal-2024-benchmarking-data, Selfalignment_10.5555/3692070.3693667, ning2024urbankgent, chang2024agentboard, ruan2024identifying}. In certain studies, GPT models have been integrated with other open-source counterparts to work collaboratively within multi-agent settings \cite{bai2025collaboration, li-etal-2023-theoryofmind, lu2024triageagent, kim2024mdagents, bo2024reflective, ghafarollahi2024protagents}. In addition to being used to benchmark and integrate other models, GPT variants are frequently used as foundational models in several agent-based systems explored in studies \cite{bai2025collaboration, jin2024large, ni2024mechagents, chen2023agentverse, ma2024coevolving, Offline_10.5555/3692070.3694566, li-etal-2023-theoryofmind, zhang-etal-2024-benchmarking-data, zhang-etal-2024-exploring, li-etal-2024-econagent, lu2024triageagent, feng-etal-2024-large, george-etal-2024-probing, yuan-etal-2025-easytool, zhu2024bootstrapping, kim2024mdagents, cheng2024sociodojo, huang2025ontheresilience, Understanding_the_Weakness_10.1145/3637528.3671650, jorgensen2024large_gpagents, cuadra2024digital, yang2024talk2care, zeng2024gesturegpt, Recommender_AI_10.1145/3731446, arumugam2025toward, huang2023benchmarking, zhang2024large, CompeteAI_10.5555/3692070.3694596, jin2024decoagent, zhang2024generative, frering2025integrating, yang2024llm, song2023llm, zhang2024large, fu2024autoguide, zhang2024privacyasst, feldt2023towards, wei2024editable, xue2025comfybench, ghafarollahi2024protagents, huang2024protchat, boiko2023autonomous, chen2025enhancing, m2024augmenting, ferber2025development}. Based on our analysis, we identified approximately 55 studies that employed GPT-4 or its variants and around 23 studies that used GPT-3.5. Beyond GPT-4 and GPT-3.5, our review also identified the use of additional OpenAI models, including Code-DaVinci-002 \cite{Describe_10.5555/3666122.3667602}, Text-DaVinci-003 \cite{song2023llm, Describe_10.5555/3666122.3667602, zhang2023large, yang2024embodied}, and GPT-2 \cite{ma2024coevolving} as agents. Furthermore, Wijk et al. used the variant o1-preview \cite{wijk2025rebench}, and o1-mini appeared in the study by Arumugam et al. \cite{arumugam2025toward}.

The Claude model family, developed by Anthropic, has emerged as a key player in the proprietary LLM domain and is often considered as a primary competitor to OpenAI models \cite{zhang2024chain, chang2024agentboard, 10802322_SMART-LLM, zhang2024privacyasst}. The Claude 3 lineup includes the Claude 3 Opus, the most advanced and capable model in the family, which has been used in several studies \cite{wu2024avatar, liu2025exploring}. Claude 3.5 and 3.7 Sonnet, recognized for their improved speed and capability, have been used as agents in several tasks such as LLM behavior elicitation, time-series ML engineering, and comparing research and development (R\&D) capabilities \cite{wijk2025rebench, cai2025llm, li2025eliciting}. Earlier generation models, such as Claude-2, have been examined in multiple comparative evaluation studies \cite{zhang-etal-2024-exploring, ruan2024identifying}, and the research capabilities of agents powered by Claude-1 have been benchmarked \cite{huang2023benchmarking}. Moreover, GPT models excel in reasoning, instruction-following, and tool integration, making them highly versatile and widely adopted for general-purpose and multi-agent applications. However, Claude 3 models, especially Opus and 3.5/3.7 Sonnet, provide faster responses and stronger performance in R\&D.

\begin{table*}[ht!]
	\centering
	\caption{Overview of the open-source LLMs for LLM agent research}
	\begin{scriptsize}
		\begin{tabular}{p{3.3cm}p{5cm}p{4cm}p{5cm}}
			\midrule
			\textbf{Model Family} & \textbf{Model Variants} & \textbf{Primary Applications} & \textbf{Key Differences} \\
			\midrule
			Meta LLaMA \cite{zhang2024building,zheng2024steveeye,wu2024chateda,xie2024can,ning2024urbankgent,jin2024large,qin2024toolllm,ma2024coevolving,yin-etal-2024-agentlumos,feng-etal-2024-large,schumann2024velma,Recommender_AI_10.1145/3731446,pang2024kalm,yang2024watch,xu2024large,Understanding_the_Weakness_10.1145/3637528.3671650,xu2024lemur,xia2024generation,hemken-etal-2025-large,singh2024personal,liu2025drbioright,yuan-etal-2025-easytool,qiao2024agent,chang2024agentboard,huang2025ontheresilience,goodell2025large,klein2025fleet,zhang-etal-2024-benchmarking-data} & LLaMA 2 (7B, 13B, 70B), LLaMA 3 (8B, 70B), LLaMA-3.1 (11B), LLaMA-3.2 (90B), CodeLLaMA (7B, 34B), ToolLLaMA-7B, COLLaMA-2 & Zero-shot eval, LoRA fine-tuning, tool use, programming agents, collaborative agents & Open-weights, broad sizes, strong coding, huge ecosystem, multilingual, tool-use \\
			Mistral AI \cite{jin2024large,zhang-etal-2024-timearena,qian-etal-2024-tellmemore,wang-etal-2024-sotopia,abdaljalil-etal-2025-theorem,qiao2024agent,zhai2024fine,liu2025exploring} & Mistral-7B, Mixtral-8×7B & Efficient inference, strong generation quality, GPT-3.5 comparable coding & Inference-efficient, Mixture of Experts (MoE), fast throughput, long-context, strong coding \\
			Google Gemma \cite{zhou2024star,qiao2024agent,tennant2025moral} & Gemma-7B, Gemma-2 (2B) & Retrieval-augmented planning, low-resource numerical agents & Lightweight, safety-tuned, efficient fine-tuning \\
			Alibaba Qwen \cite{li2025llm,zhou2024star} & Qwen-max, Qwen 2.5 (72B) & Star-Agents, SMAC reinforcement & Chinese-English, long-context, tool-use \& enterprise features, multimodal \\
			DeepSeek AI \cite{li2025llm,ColaCare_10.1145/3696410.3714877,abdaljalil-etal-2025-theorem} & DeepSeek-r1-70B, DeepSeek-7B, DeepSeek V2.5 & Open-source agent models & Reasoning-focused, math/code strength, cost-efficient, RL-style alignment, long-context \\
			Vicuna/WizardLM \cite{zhang-etal-2024-timearena,Selfalignment_10.5555/3692070.3693667,RCAgent_10.1145/3627673.3680016,ruan2024identifying,xie2024can,Reason_for_future_10.5555/3692070.3693331,de2025multimodal,xia2024generation} & Vicuna-13B, Wizard-Vicuna-30B, WizardLM-70B & Conversational agents, high-capacity chat models & Instruction-tuned chat, easy fine-tune \\
			Zhipu AI \cite{zhou2024star} & GLM-4, ChatGLM3 & General chat and task execution & Chinese-first bilingual, tool-use \& agents, long-context, enterprise stack \\
			Microsoft \cite{li2025llm} & Phi-2, Phi-3.5 Mini & Compact inference agents & Ultra-small, on-device ready, textbook-style data curation, inference-efficient \\
			OpenChat/Baichuan \cite{ruan2024identifying} & OpenChat-3.5, Baichuan-13B & Dialogue-optimized agents & Conversational alignment, Chinese-English, open-weights \\
			LongChat \cite{bo2024reflective} & LongChat-7B & Long-context critic agents & Extra-long context, memory retention, position scaling tricks \\
			Multimodal \cite{zhai2024fine,wang2024omnijarvis,hong2024cogagent} & LLaVA-7B, LLaVA-v1.6-mistral-7B, CogVLM-17B & Vision-language agents, GUI-based interactions & Vision-language, image grounding, visual question-answering \& captioning, perceptual reasoning \\
			Generative/Visual \cite{zheng2024steveeye} & CLIP, Stable Diffusion XL, VQ-GAN & Image generation, visual embedding for agents & Image generation, latent diffusion, visual embeddings \\
			\midrule
		\end{tabular}
	\end{scriptsize}
	\label{tab:open_source_llm_agents}
    \vspace{-10pt}
\end{table*}

In LLM agent research, Google’s proprietary models are frequently featured, particularly in frameworks involving multiple model evaluations. The Gemini series is the most prominent \cite{zhang2024privacyasst, barra2025prompt}, with Gemini Pro emerging as the version most often used in agent development and evaluation \cite{zhang-etal-2024-timearena, zhang-etal-2024-benchmarking-data, kim2024mdagents, behari2024decision}. Gemini 1.5 Pro, known for its large context window, has been used in LLM-based agent reasoning under distraction tasks \cite{george-etal-2024-probing}, and Gemini 1.0 Pro can be found in specific LLM agent benchmark studies \cite{liu2025exploring, zhang2024chain}. The PaLM lineup is also represented, and PaLM 2 appears in the work of Zhang et al. \cite{zhang2024chain}. Other significant models include FLAN-T5, which is noted as a foundational component for enterprise-level agents \cite{muthusamy-etal-2023-towards}. In addition to proprietary models from well-known tech companies, the Mistral medium-2312 \cite{liu2025exploring}, Chinchilla-LM (70B) \cite{lampinen2023passive}, and HyperCLOVA (82B) \cite{Understanding_10.1145/3544548.3581503} models are also included in our review of LLM agents.

\subsection{Open-sourced LLMs for agentic applications}
The landscape of open-source LLMs used in agent research and development is rapidly diversifying, offering researchers and developers powerful alternatives to proprietary models \cite{Recommender_AI_10.1145/3731446, xu2024lemur}. With models like Meta's LLaMA, Mistral, Google's Gemma, and the Qwen model from Alibaba gaining traction, open models now support a wide spectrum of capabilities, from code generation and dialogue to visual reasoning and collaborative decision making \cite{feng-etal-2024-large, yuan-etal-2025-easytool, schumann2024velma, Understanding_the_Weakness_10.1145/3637528.3671650}. An overview of all open source LLMs studied is presented in Table \ref{tab:open_source_llm_agents}.

Meta’s LLaMA suite of models is widely regarded as the default open source platform that supports contemporary research on LLM agents. The LLaMA 2 series \cite{zhang2024building, zheng2024steveeye, wu2024chateda, xie2024can, ning2024urbankgent}, comprising models with 7B \cite{jin2024large, qin2024toolllm, ma2024coevolving, yin-etal-2024-agentlumos, feng-etal-2024-large, schumann2024velma, Recommender_AI_10.1145/3731446, pang2024kalm, yang2024watch, xu2024large}, 13B \cite{yin-etal-2024-agentlumos, feng-etal-2024-large, Understanding_the_Weakness_10.1145/3637528.3671650}, and 70B \cite{Understanding_the_Weakness_10.1145/3637528.3671650, xu2024lemur, xia2024generation} parameters, remains integral to the field, frequently used for zero-shot assessments via chat-tuned versions or as foundational checkpoints for lightweight LoRA alignment. With the advent of LLaMA 3 \cite{hemken-etal-2025-large, ning2024urbankgent, singh2024personal, liu2025drbioright}, this foundation has grown significantly: its 8B \cite{singh2024personal, yuan-etal-2025-easytool, qiao2024agent, chang2024agentboard} and 70B \cite{yuan-etal-2025-easytool, huang2025ontheresilience,goodell2025large} instruction-tuned models are now integral to planning, tool-use, and multi-agent benchmarks. Additional advancements include the 'LLaMA-3-8B-instruct-8k' for longer input contexts, as well as newer model checkpoints such as LLaMA-3.1 (11B) \cite{klein2025fleet} and LLaMA-3.2 (90B) \cite{klein2025fleet}, which support more advanced reasoning. Additionally, several specialized extensions have emerged alongside the base models: CodeLLaMA 7B and 34B enable programming agents \cite{zhang-etal-2024-benchmarking-data}; ToolLLaMA-7B facilitates the generation of structured tool calls \cite{yuan-etal-2025-easytool}; and COLLaMA-2 \cite{zhang2024building} is designed for collaborative behaviors in AI embodied systems. These derivatives are integrated as either immutable inference modules prompted externally or as adaptable architectures through adapter-based tuning.

The models developed by Mistral AI offer a highly efficient line of alternatives, optimized for strong performance at relatively moderate computational cost. The dense Mistral-7B model is frequently preferred \cite{jin2024large, zhang-etal-2024-timearena, qian-etal-2024-tellmemore, wang-etal-2024-sotopia, abdaljalil-etal-2025-theorem, qiao2024agent, zhai2024fine} over LLaMA-2-7B, delivering better generation quality. On the other hand, the Mixtral-8×7B \cite{zhang-etal-2024-timearena, liu2025exploring} model, also known as open-mixtral-8×7B, utilizes a sparse expert architecture to achieve performance similar to GPT-3.5 in coding and planning tasks, while still benefiting from light inference.

The Gemma model family from Google represents a valuable addition to the open source LLM space \cite{zhou2024star}, the 7B version is already being integrated into retrieval-augmented planning architectures \cite{qiao2024agent}. Furthermore, the Gemma-2 (2B) model tuned according to the instruction is utilized in studies \cite{tennant2025moral} involving numerical agents operating in low-resource or constraint environments.

A wider collection of open-source models contributes further diversity to the landscape of agent baselines. Alibaba’s Qwen line (Qwen-max and the 72B parameter Qwen 2.5) competes strongly in Star Agents and SMAC reinforcement tasks \cite{li2025llm, zhou2024star}; DeepSeek AI’s DeepSeek-r1-70B, DeepSeek-7B, and DeepSeek V2.5 achieve a top-tier score among open source models \cite{li2025llm, ColaCare_10.1145/3696410.3714877, abdaljalil-etal-2025-theorem}. Conversation-centric agents often adopt Vicuna-13B or the larger Wizard-Vicuna-Uncensored 30B \cite{zhang-etal-2024-timearena, Selfalignment_10.5555/3692070.3693667, RCAgent_10.1145/3627673.3680016, ruan2024identifying, xie2024can, Reason_for_future_10.5555/3692070.3693331, de2025multimodal}, while the instruction-tuned WizardLM 70B serves as a high-capacity but fully open chat baseline \cite{xia2024generation}.  GLM-4 and ChatGLM3 (Zhipu AI), Microsoft’s compact Phi-2 and Phi-3.5 Mini, the OpenChat-3.5 chat model optimized for dialogue, Baichuan's 13B chat model, and the LongChat-7B enhanced by long context, the latter fine-tuned as a reflective critic in collaborative frameworks \cite{zhou2024star, li2025llm, abdaljalil-etal-2025-theorem, bo2024reflective, zhang-etal-2024-timearena}.

\begin{table*}[ht!]
\centering
\caption{Overview of the tools usage across LLM agents' capabilities}
\begin{scriptsize}
\renewcommand{\arraystretch}{1.3}
\begin{tabular}{p{3cm}p{3.3cm}p{2.4cm}p{4.1cm}p{4.2cm}}
    \toprule
    
    \textbf{Domain} & \textbf{Single-Agent} & \textbf{Multi-Agent} & \textbf{(Single + Multi)} & \textbf{Key Differences} \\
    \midrule
    Interactive and Embodied Environments \cite{zhu2024bootstrapping,frering2025integrating,zheng2024steveeye,Describe_10.5555/3666122.3667602,jorgensen2024large_gpagents,shinn2023reflexion,li2025llm,zhang2024proagent,jin2024decoagent} &
    MineDojo (Minecraft), MarioAI, ALFWorld &
    SMAC, Overcooked-AI, DeCoAgent (Smart Contracts) &
    AI2THOR, ROS 2, Gazebo, Clearpath Husky &
    Open-source stacks; simulation \& robotics; embodied/multimodal interaction; native multi-agent tasks; RL \& perception–action loops  \\
    Code Generation, API Use, and System-Level Integration \cite{chen2023agentverse,zhang-etal-2024-benchmarking-data,ni2024mechagents,lu2024triageagent,ghafarollahi2024protagents,chen2025enhancing,qian-etal-2024-tellmemore,ramos2025review,qin2024toolllm,wu2024chateda,cuadra2024digital} &
    Code Interpreters, Copilot, Excel, Power BI, Jupyter AI, Chapyter, CoML, RDKit, Scikit-learn, ChatEDA (OpenROAD), Speechly &
    AutoGen &
    RapidAPI &
    Open-source \& proprietary; code execution; API orchestration; system/engineering integration \\
    Knowledge Grounding, Web Search, and Structured Retrieval \cite{chen2023agentverse,qian-etal-2024-tellmemore,yin-etal-2024-agentlumos,cheng2024sociodojo,lu2024triageagent,feng-etal-2024-large,yuan-etal-2025-easytool,kim2024mdagents,m2024augmenting} &
    --- &
    --- &
    Bing Search API, Google Search, DuckDuckGo, Wikipedia API, PubMed, UMLS, ESI Handbook, FinBERT, MedRAG, ChemCrow (RoboRXN) &
    Open-source \& proprietary mix; textual/structured retrieval; domain-specialized; RAG-ready \\
    \bottomrule
\end{tabular}
\end{scriptsize}
\label{tab:tool_usage_capabilities}
\vspace{-10pt}
\end{table*}

This domain is further enriched by multimodal and architectural developments. Vision language assistants such as LLaVA-7B and the more recent LLaVA-v1.6-mistral-7B combine a CLIP-style image encoder with a chat-based LLM, allowing agent capabilities in embodied environments and GUI-based interfaces \cite{zhai2024fine, wang2024omnijarvis}. CogVLM-17B expands this functionality with enhanced visual reasoning and understanding capacity \cite{hong2024cogagent}. In the realms of perception and generative synthesis tasks, agents commonly use OpenAI CLIP, Stable Diffusion XL for high-resolution image generation, and VQ-GAN to produce latent image tokenization \cite{zheng2024steveeye}. These components are often combined with the LLaMA-2 models in interactive applications. These model ecosystems establish the foundational open source landscape on which LLM agent research is currently based.

A critical comparison of proprietary versus open-source models reveals divergent evolutionary trajectories in LLM agent research. Although proprietary models such as GPT-4 and Claude demonstrate superior performance on complex reasoning benchmarks, their widespread deployment involves notable trade-offs. First, the cost of inference remains prohibitively high for large-scale or real-time applications. Second, the closed-source nature of GPT models restricts model transparency and limits fine-tuning flexibility compared to open-source alternatives. Furthermore, reliance on cloud-based APIs raises privacy and security concerns for sensitive domains such as healthcare and finance, where data cannot be externally shared. Recent studies also indicate that domain-specific tasks often require task-adaptive fine-tuning to achieve competitive accuracy. In this regard, open-source models such as Mistral-7B, DeepSeek-V2.5, and LLaMA-2-70B, when fine-tuned effectively, can achieve between 60\% and 95\% of GPT-4 and other models’ performance while being significantly more cost-efficient \cite{wang-etal-2024-sotopia, qian-etal-2024-tellmemore, ColaCare_10.1145/3696410.3714877, qin2024toolllm}. As a result, hybrid model architectures are increasingly adopted, where proprietary models are used as a high-level supervisor for data generation or validation, while smaller open-source models manage routine inference and domain-specific downstream tasks.

\section{External tool integration across LLM agent workflows}
\label{5_external_tool}

The transformation of an LLM into an autonomous agent fundamentally relies on its capacity to engage with external systems and sources beyond the operational reach of its pre-trained data \cite{jin2024large, zhang2024generative, CHEN2025151, jin2024decoagent, ramos2025review}. It is facilitated through an increasingly rich ecosystem of external tools, APIs, and software frameworks \cite{chen2023agentverse, qin2024toolllm}. Far from being optional add-ons, the integration of these tools serves as foundational components that empower agents to access real-time information, perform complex operations, and the ability to interact with environments ranging from software platforms to physical systems \cite{li2025llm, jin2024large, ni2024mechagents, m2024augmenting}. A summary of the tools covered in various studies is presented in Table \ref{tab:tool_usage_capabilities}. Our review of existing studies identifies several distinct patterns in tool utilization, from basic retrieval tasks to the management of sophisticated multi-agent systems.

\subsection{Usage of tools across knowledge grounding, web search, and structured retrieval}

One of the fundamental applications of tools is to mitigate the limitations of an LLM’s fixed internal knowledge. These tools enable agents to retrieve real-time data and access domain-specific specialized knowledge repositories \cite{lu2024triageagent, kim2024mdagents, ColaCare_10.1145/3696410.3714877, zhang2024generative, ferber2025development}.

The integration of web search APIs is the most common and standard strategy for equipping agents with real-time data retrieval capabilities and overcoming the limitations of LLM knowledge. Studies frequently use tools such as the Bing Search API, Google Search, and DuckDuckGo to extend agents’ access to up-to-date online content \cite{chen2023agentverse, qian-etal-2024-tellmemore, yin-etal-2024-agentlumos, cheng2024sociodojo}. 

In addition, these tools are often paired with structured external knowledge sources. In particular, agents operate in domain-specific settings. For example, biomedical agents are linked to PubMed and UMLS knowledge bases, one also found using the Emergency Severity Index (ESI) handbook \cite{lu2024triageagent} for access to specialized medical information, FinBERT to classify sentiment of financial texts, while the Wikipedia Web API is a widely used source for general encyclopedic information \cite{feng-etal-2024-large, yin-etal-2024-agentlumos, lu2024triageagent, cheng2024sociodojo, yuan-etal-2025-easytool}. 

Frameworks such as MedRAG apply a retrieval augmented generation (RAG) approach explicitly to ensure that medical agents provide responses that are factually grounded in validated clinical knowledge \cite{kim2024mdagents}. Bran et al. \cite{m2024augmenting} introduced ChemCrow to overcome the limitations of LLM in chemistry, which leverages GPT-4, RoboRXN, and 18 expert tools to autonomously plan and execute intricate chemical tasks in organic synthesis, drug discovery, and materials design \cite{m2024augmenting}.

\subsection{Usage of tools across code generation, API use, and system-level integration}
The generation and execution of code serves as a key mechanism of agentic behavior, empowering agents to perform sophisticated calculations, handle complex data operations, and engage with various software environments. 

A widely used tool in agentic systems is the diverse types of Code Interpreter developed in a number of studies \cite{chen2023agentverse, zhang-etal-2024-benchmarking-data}. This is particularly relevant for data science agents, which are evaluated for their integration with platforms and tools such as Excel, Copilot, Power BI, Jupyter AI, Chapyter, and CoML \cite{zhang-etal-2024-benchmarking-data}. Some studies also incorporate AutoGen, a framework designed to build agentic AI systems through multi-agent interactions and improved LLM inference capabilities \cite{ni2024mechagents, lu2024triageagent, ghafarollahi2024protagents, chen2025enhancing}. These agents often produce code to interact with file systems \cite{qian-etal-2024-tellmemore}, command-line utilities \cite{qian-etal-2024-tellmemore}, and domain-specific libraries, such as RDKit for chemical informatics \cite{ramos2025review} and Scikit-learn for machine learning. Qin et al. found that using ChatGPT's function calling mechanism, LLMs can interact with 16,464 RESTful APIs sourced from RapidAPI by utilizing a neural API retriever and structured API documentation \cite{qin2024toolllm}. 

In the EDA sector, the ChatEDA agent demonstrates this capability through its Python wrapper-based interaction with the OpenROAD platform \cite{wu2024chateda}. Cuadra et al., in their study, used Speechly to support the speech-to-text task for the purpose of entering health data \cite{cuadra2024digital}.

\subsection{Usage of tools across interactive and embodied environments}
To investigate complex and interactive behaviors, agents are often employed in simulated or real-world environments. They function as tools that offer detailed, state-dependent feedback in response to agent interactions. Zhu et al. \cite{zhu2024bootstrapping} employed AI2THOR, a simulation platform that supports realistic 3D environments with physical interactions and dynamic visual states, for the purpose of assessing embodied agents within household environments. 

In human-robot interaction, Frering et al. \cite{frering2025integrating} utilized ROS 2 and the Gazebo simulator with a model like the Clearpath Husky. Interactive environments used in agentic research include virtual platforms like game worlds such as Minecraft (via MineDojo) \cite{zheng2024steveeye, Describe_10.5555/3666122.3667602}, the MarioAI platform \cite{jorgensen2024large_gpagents}, and text-based simulations such as ALFWorld \cite{shinn2023reflexion}. In multi-agent contexts, popular benchmarks include the StarCraft Multi-Agent Challenge (SMAC) and the collaborative game Overcooked-AI \cite{li2025llm, zhang2024proagent}. An especially innovative example can be seen in the DeCoAgent framework, where LLM agents autonomously coordinate with the help of smart contracts, overcoming limitations of static, closed multi-agent environments \cite{jin2024decoagent}.

\section{Frameworks for building LLM agents}
\label{6_framework4agents} 
LLMs align closely with the core properties of agents in AI, making them strong candidates for agent foundations. First, they demonstrate autonomy by performing tasks without granular instructions, adapting responses to input, and generating creative content independently \cite{21,17,98}. In terms of reactivity, LLMs can now handle multimodal inputs and interact with their environment using embodiment and tool integration, despite the latency caused by the textual reasoning stages \cite{schick2023toolformer, qin2024tool, driess2023palm, kang2024thinkactdecisiontransformers}. Their proactivity is seen in their ability to reason and plan when asked, including setting goals and decomposition of tasks in dynamic settings \cite{xi2025llmagentsurvey}. 

This section systematically reviews the frameworks developed for building LLM agents, highlighting their architectural designs, core capabilities, and operational domains. Categorizing existing solutions and analyzing their design principles provides a foundational lens for understanding how LLM agents are constructed and deployed in diverse scenarios.

\subsection{Basic architecture of an LLM agent}
Building LLM-based agents requires a systematic architectural design that enables LLMs to interact autonomously with their environment, recall relevant information, plan strategically, and execute appropriate actions. Unlike traditional question-answering models, these agents continuously perceive, reason, and adapt to various tasks. A widely adopted architecture for LLM agents primarily includes four core components: profile definition, memory, planning, and action execution, which together form a feedback-driven system in which memory shapes planning, actions modify memory, and update the agent's operational profile.

\vspace{0.3em} \noindent \textbf{\textit{Profiling.}}
The profiling module defines an agent's operational persona (e.g., developer, advisor, or task-specific role), conditioning its behavior and role policy through static profiles defined by experts or dynamic generative mechanisms \cite{hong2023metagpt, 10.1145/3672459}. Static profiles encode domain knowledge and role-specific behaviors, while dynamic profiles simulate human-like variability by generating diverse agent personas through prompt engineering or parameter sampling \cite{UserBehavior_10.1145/3708985}. These profiles may include demographic data, personality traits, and social relationships, significantly influencing downstream decisions in memory retrieval, planning strategies, and action selection.

\vspace{0.3em} \noindent \textbf{\textit{Memory.}}  
Memory enables agents to maintain context across interactions through short-term (prompt-based) and long-term (externally stored) forms. Short-term memory supports in-the-moment reasoning by storing dialogue history and environmental signals, but is limited by LLM context windows \cite{wang2024llmautonomous}. Long-term memory captures reusable skills, patterns, or tools from past interactions. Memory formats vary from natural language, embeddings, databases (e.g., SQL), to structured lists, each chosen based on task requirements. Common operations include reading (prioritized by recency, relevance, and importance), writing (handling duplication and overflow), and reflection (summarizing past experiences into high-level insights).

\vspace{0.3em} \noindent \textbf{\textit{Planning.}} Planning modules decompose complex tasks using strategies like CoT \cite{10.5555/3600270.3602070} and Tree-of-Thought (ToT) \cite{NEURIPS2023_271db992}. Planning can occur with or without feedback. Feedback-free strategies often use stepwise prompting (e.g., CoT, ToT). Feedback-based iteration allows agents to dynamically adapt plans using signals from the environment, human input, or memory reflections.

\vspace{0.3em} \noindent \textbf{\textit{Action Execution.}}
The action module translates plans into executable outputs. Actions may follow retrieved memories or precomputed plans. Furthermore, agents may engage in feedback loops in which the outcome of an action informs subsequent memory updates, plan revisions, or behavioral adaptations and can impact both the environment and the agent itself \cite{luo2025largelanguagemodelagent}. Moreover, this module is crucial for grounding the agent in real or simulated environments.

Together, these components enable LLMs to function as autonomous agents, reasoning, remembering, planning, and acting in open-ended and evolving tasks. This modular architecture is foundational to both single-agent and multi-agent systems discussed in Section~\ref{sec:plan_reason}.

\subsubsection{Single-agent LLM system}
\begin{figure}[!ht]
\centering
\includegraphics[scale=0.38]{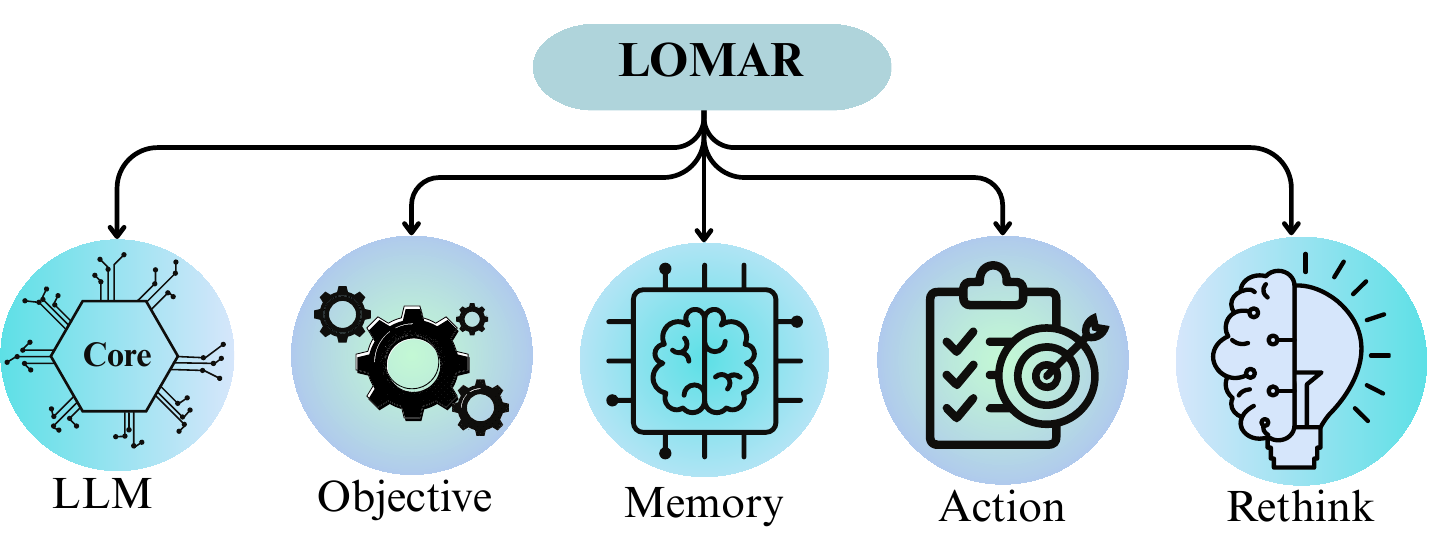} 
\caption{An illustration of the LOMAR framework}
\label{fig:lomar}
\end{figure}

Single-agent LLM frameworks are often superior in generalization, reasoning, and task execution without additional model training. A single agent LLM can be conceptualized using a five-core component (LOMAR) \cite{cheng2024exploringlargelanguagemodel}: \textbf{L}LM, \textbf{O}bjective, \textbf{M}emory, \textbf{A}ction, and \textbf{R}ethink. Figure~\ref{fig:lomar} illustrates the LOMAR framework.

\textbf{LLM} is the core of an agent, which allows task planning and decision making based on current inputs, memory, and feedback \cite{liu2024agentbench}. In addition, a \textbf{objective} defines the target goal and guides how the agent breaks down complex tasks to formulate a strategy. \textbf{Memory} is a dynamic storehouse of past interactions and relevant contextual information, enabling the agent to adapt its behavior thoughtfully \cite{wang2023voyager, xu2025amemagenticmemoryllm}. \textbf{Action} refers to the operational capabilities of the agent. It executes commands, interacts with tools, or communicates outputs \cite{schick2023toolformer}; and \textbf{Rethink} facilitates reflective learning by evaluating previous actions and environmental feedback to inform future decisions \cite{yao2023react}.

These agents are designed to learn from continuous interactions and maintain coherent behavior over time through memory systems. Unlike multi-agent architectures that depend on collaborative dynamics, single-agent systems operate autonomously to complete tasks independently.

\begin{figure}[ht!]
    \centering
    \includegraphics[scale=0.4]{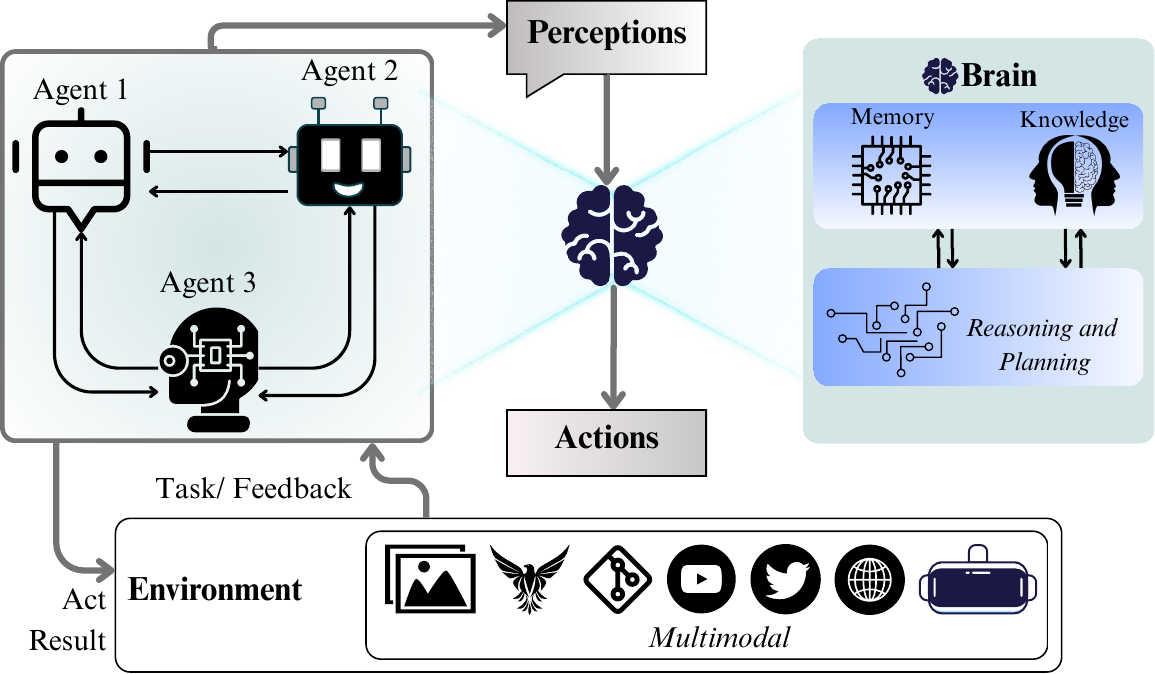} 
    \caption{A general overview of a multi-agent LLM system. Here, three agents operate within a multimodal environment where they act, generate results, and exchange feedback. Each agent is equipped with internal modules (brain), including memory, reasoning, and planning, that guide its behavior. Through collaborative communication, agents perceive the environment, coordinate strategies, and ultimately take action.} 
    \label{fig:multiagent}
\end{figure}
\subsubsection{Multi-agent LLM systems}

Multi-agent LLM systems are designed for coordinated collaboration, where multiple agents communicate, adapt, and solve problems together (Figure~\ref{fig:multiagent}). A description of how multiple agents observe, interact with, and adjust to different environments to work together toward shared objectives is given to highlight the foundation for effective multi-agent collaboration.

\vspace{0.3em} \noindent \textbf{\textit{Agents-environment interaction.}} In a multi-agent LLM framework, individual agents operate within simulated, physical, or abstract environments based on their assigned roles. This layer allows agents to detect contextual changes, perform environment-specific actions, and adapt dynamically in response to feedback.

\vspace{0.3em} \noindent \textbf{\textit{Communication structures.}} In this layer, three major paradigms are prevalent for agent communication, including \textit{cooperative} (to work together toward shared goals), \textit{debate} (to argue opposing views to reach consensus), and \textit{competitive} (to pursue individual objectives). Communication can be centralized, decentralized, or shared memory-based, where agents publish and subscribe to a shared message pool.

\vspace{0.3em} \noindent \textbf{\textit{Adaptive learning through feedback.}} Agents learn and adapt based on the feedback of the environment and interactions with other agents and humans. They can refine agent behavior by adjusting their strategies in response to dialogues or integrating human corrections, enabling flexible learning across dynamic scenarios.

\vspace{0.3em} \noindent \textbf{\textit{Dynamic agent role.}} Multi-agent LLM frameworks support real-time agent generation or profile updates. This includes generating new agents with targeted roles or modifying goals in mid-task. However, as agent populations grow, maintaining coordination becomes increasingly vital for overall system performance.

\subsection{Common LLM agent frameworks}
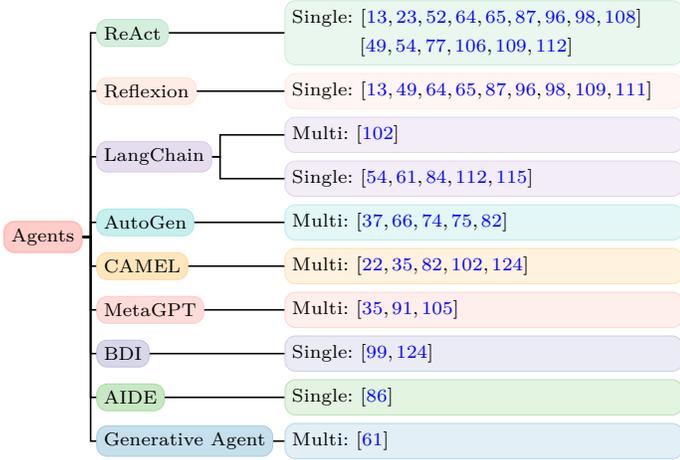
\begin{figure}[ht]
\centering
\scriptsize
\setlength{\lineskip}{0pt}
\setlength{\baselineskip}{0.7\baselineskip}

\definecolor{softblue}{RGB}{158, 202, 225}
\definecolor{softgreen}{RGB}{161, 217, 155}
\definecolor{softpurple}{RGB}{188, 189, 220}
\definecolor{softpink}{RGB}{252, 197, 192}
\definecolor{softorange}{RGB}{255, 213, 145}
\definecolor{softcyan}{RGB}{162, 228, 229}
\definecolor{softlavender}{RGB}{213, 196, 228}
\definecolor{softpeach}{RGB}{255, 223, 211}
\definecolor{softmint}{RGB}{183, 228, 199}
\definecolor{rootcolor}{RGB}{255, 183, 178}

\begin{forest}
  for tree={
    grow=0,
    child anchor=west,
    parent anchor=east,
    anchor=west,
    l sep=1.5mm,
    s sep=1mm,
    tier/.option=level,
    edge path={
      \noexpand\path [draw, \forestoption{edge}, line width=0.6pt]
      (!u.parent anchor) -- +(3pt,0) |- (.child anchor)\forestoption{edge label};
    },
    fit=band,
    rounded corners,
    font=\rmfamily,
  }
[Agents, anchor=center, fill=rootcolor!70, draw=rootcolor!90, text=black, line width=0.8pt
  [Generative Agent, fill=softblue!60, draw=softblue!90, text=black
    [Multi: \cite{zhang2024generative}, align=left, text width=5.1cm, fill=softblue!30, draw=softblue!70]
  ]
  [AIDE, fill=softgreen!60, draw=softgreen!90, text=black
    [Single: \cite{wijk2025rebench}, align=left, text width=5.1cm, fill=softgreen!30, draw=softgreen!70]
  ]
  [BDI, fill=softpurple!60, draw=softpurple!90, text=black
    [Single: \cite{frering2025integrating, xie2024can}, align=left, text width=5.1cm, fill=softpurple!30, draw=softpurple!70]
  ]
  [MetaGPT, fill=softpink!60, draw=softpink!90, text=black
    [Multi: \cite{zhang-etal-2024-benchmarking-data, huang2025ontheresilience, klein2025fleet}, align=left, text width=5.1cm, fill=softpink!30, draw=softpink!70]
  ]
  [CAMEL, fill=softorange!60, draw=softorange!90, text=black
    [Multi: \cite{klein2025fleet, CompeteAI_10.5555/3692070.3694596, li2023camel, UserBehavior_10.1145/3708985, xie2024can}, align=left, text width=5.1cm, fill=softorange!30, draw=softorange!70]
  ]
  [AutoGen, fill=softcyan!60, draw=softcyan!90, text=black
    [Multi: \cite{lu2024triageagent ,CompeteAI_10.5555/3692070.3694596, ghafarollahi2024protagents,huang2024protchat,chen2025enhancing}, align=left, text width=5.1cm, fill=softcyan!30, draw=softcyan!70]
  ]
  [LangChain, fill=softlavender!60, draw=softlavender!90, text=black
    [Single: \cite{Recommender_AI_10.1145/3731446, huang2023benchmarking, feldt2023towards, zhang2024generative, m2024augmenting}, align=left, text width=5.1cm, fill=softlavender!30, draw=softlavender!70]
    [Multi: \cite{UserBehavior_10.1145/3708985}, align=left, text width=5.1cm, fill=softlavender!30, draw=softlavender!70]
  ]
  [Reflexion, fill=softpeach!60, draw=softpeach!90, text=black
    [Single: \cite{zhang-etal-2024-timearena, singh2024personal, feng-etal-2024-large, Understanding_the_Weakness_10.1145/3637528.3671650, arumugam2025toward, zhang2023large, shinn2023reflexion, fu2024autoguide, wu2024avatar}, align=left, text width=5.1cm, fill=softpeach!30, draw=softpeach!70]
  ]
  [ReAct, fill=softmint!60, draw=softmint!90, text=black
    [Single: \cite{shinn2023reflexion, fu2024autoguide, zhang-etal-2024-timearena, chang2024agentboard, singh2024personal, zhang2023large, Offline_10.5555/3692070.3694566, feng-etal-2024-large, RCAgent_10.1145/3627673.3680016} \vspace{1ex}\\ \hspace{7.6ex} \cite{ruan2024identifying, wu2024avatar, Understanding_the_Weakness_10.1145/3637528.3671650, yang2024talk2care, huang2023benchmarking, m2024augmenting}, align=left, text width=5.1cm, fill=softmint!30, draw=softmint!70]
  ]
]
\end{forest}

\caption{Common LLM agent frameworks used in single- and multi-agent LLMs}
\label{fig:llm_agent_types}
\end{figure}

Several frameworks have been adopted to implement LLM agents, facilitating reasoning, decision making, memory management, and action execution in single-agent and multi-agent contexts. Figure~\ref{fig:llm_agent_types} presents a categorized overview of these frameworks based on their application in single-agent or multi-agent systems.

Among these, ReAct and Reflexion are the prominently utilized frameworks within single-agent LLM architectures. The selection between ReAct and Reflexion in LLM agents depends on the nature of the task. ReAct, by interleaving structured reasoning with action, such as planning or tool use, facilitates effective and context-aware execution. In contrast, Reflexion enables agents to retrospectively evaluate their prior outputs, learn from feedback, and refine subsequent decisions. Integrating both mechanisms allows for a balance between immediate operational efficiency and iterative self-improvement in complex tasks. AutoGen and CAMEL are widely employed in multi-agent environments. AutoGen is designed for building customizable, complex workflows that emphasize structured role coordination and task orchestration. CAMEL, on the other hand, focuses on autonomous role-playing and goal-driven collaboration, making it especially suited for dialogue-intensive or simulation-based environments. Furthermore, LangChain is also recognized as a versatile framework, applied across single-agent and even in multi-agent paradigms. While it offers a flexible, modular infrastructure for LLM applications, its relatively unopinionated design can limit its applicability for certain tasks, as it places the responsibility for key architectural decisions largely on developers.

\subsection{Domain-Specific Frameworks}

Existing research on LLM-based agents is classified according to domain-specific applications, with each domain further divided into single-agent and multi-agent architectures. We have organized the review into nine specialized domains and comprehensively analyzed the relevant works within each. 

\subsubsection{Single-Agent LLM Systems' Application Domain}

Single-agent LLM-based systems have been demonstrating remarkable performance across various application domains. These systems benefit from centralized decision making, lower communication overhead, and seamless knowledge integration, making them especially suitable for scalable and autonomous deployment. This section synthesizes evidence from recent works to highlight the breadth and depth of capabilities exhibited by single-agent LLM systems across multiple domains.

\vspace{0.3em} \noindent \textbf{\textit{Healthcare.}} LLM agents in healthcare have shown promise in tasks ranging from personalized consultation to domain-specific scientific analysis. Conversational agents provide personalized health advice and support elderly care through natural interfaces \cite{yang2024talk2care, abbasian2025conversational}. Others optimize public health policies using reinforcement learning (RL) \cite{behari2024decision} or operate in real-world clinical settings \cite{Understanding_10.1145/3544548.3581503}. More specialized agents help with cancer protein analysis and biomedical research tasks \cite{liu2025drbioright}, underscoring the ability of single agent systems to operate in high-stakes and knowledge-intensive settings.

\vspace{0.3em} \noindent \textbf{\textit{Software engineering.}} It is one of the major areas where single-agent LLM systems offer substantial automation and support. They improve code generation \cite{bai2025collaboration}, perform automated software testing \cite{feldt2023towards}, and diagnose complex failures in cloud systems through the use of integrated tools \cite{RCAgent_10.1145/3627673.3680016}. These agents operate autonomously in mobile and distributed environments, even as the limitations in edge cases are explored \cite{Understanding_the_Weakness_10.1145/3637528.3671650}, showcasing their adaptability and evolving maturity.

\vspace{0.3em} \noindent \textbf{\textit{Scientific research.}} LLMs are increasingly being positioned as autonomous scientific researchers. Agents can benchmark themselves against human researchers in AI tasks \cite{wijk2025rebench}, conduct chemistry experiments autonomously \cite{boiko2023autonomous}, or aid in semiconductor and biological sequence analysis \cite{zhang2024largeSC, de2025multimodal}. These systems highlight how single-agent architectures enable sustained domain-specific scientific inquiry without constant human intervention \cite{huang2023benchmarking}.

\vspace{0.3em} \noindent \textbf{\textit{Robotics.}} In robotics, single-agent LLM systems are cognitive controllers capable of grounded reasoning and planning. Integrating BDI architectures \cite{frering2025integrating}, visual grounding \cite{zheng2024steveeye, frering2025integrating}, and few-shot learning for robot task planning \cite{song2023llm}, these agents demonstrate their ability to translate high-level instructions into actions in physical or simulated environments. Embodied LLM agents are further empowered by cognitive initialization and explainability features \cite{zhu2024bootstrapping}, pushing forward the boundary of autonomous, language-guided robotics.

\vspace{0.3em} \noindent \textbf{\textit{Recommendation system.}} Conversational LLM agents are deployed in recommendation systems to improve personalization and interaction. Agents simulate realistic user behavior for testing \cite{UserBehavior_10.1145/3708985} and provide interactive recommendation dialogues \cite {Recommender_AI_10.1145/3731446}, capturing evolving user intent with minimal supervision.

\vspace{0.3em} \noindent \textbf{\textit{Urban systems.}} In urban systems, LLM agents contribute to intelligent infrastructure management. They optimize mixed vehicle parking strategies \cite{jin2024large}, generate comprehensive urban knowledge graphs \cite{ning2024urbankgent}, and facilitate vision- and language-based street navigation \cite{schumann2024velma}. These use cases highlight the scalability of single-agent systems in real-world multimodal environments.

\vspace{0.3em} \noindent \textbf{\textit{General-purpose systems.}} A growing body of work seeks to develop general-purpose agent frameworks that empower LLMs with reasoning, planning, and tool use capabilities \cite{qin2024toolllm, Offline_10.5555/3692070.3694566, qian-etal-2024-tellmemore, yuan-etal-2025-easytool, xu2024lemur}. These systems support multitask environments \cite{zhang-etal-2024-timearena}, enhance collaboration with humans \cite{feng-etal-2024-large}, and enable long-horizon planning through world knowledge and instruction tuning \cite{qiao2024agent, fu2024autoguide}. Several benchmarks have been proposed to assess performance in data science \cite{zhang-etal-2024-benchmarking-data}, safety \cite{ruan2024identifying}, and reasoning tasks \cite{abdaljalil-etal-2025-theorem, liu2024agentbench}, while architectures address self-improvement \cite{Selfalignment_10.5555/3692070.3693667} and causal reasoning \cite{lampinen2023passive}. RL paradigms adapted to LLM further boost agent autonomy \cite{zhang2023large, shinn2023reflexion, pang2024kalm}.

Beyond standard domains, single-agent LLM systems have also been adapted for specialized applications and use cases such as hand gesture understanding \cite{zeng2024gesturegpt} and time-series engineering challenges \cite{cai2025llm}. These illustrate how adaptable the agentic paradigm is to domain constraints and modality fusion.

\vspace{0.3em} \noindent \textbf{\textit{Security and privacy.}} Given their growing autonomy, security and privacy are critical concerns for LLM agents. Many studies have discovered backdoor vulnerabilities \cite{yang2024watch}, privacy risks during tool use \cite{zhang2024privacyasst}, and the challenge of enforcing confidentiality and intentional alignment \cite{hemken-etal-2025-large, khan2025in}. 

These efforts highlight the need for principled design and evaluation in real-world deployments. Agents have to act ethically and socially responsibly. Work in the domain of social intelligence and ethics ensures that LLM agents adhere to the principles of moral reasoning and avoid harmful behaviors \cite{tennant2025moral}, demonstrating the growing maturity in aligning LLM actions with human values.

\subsubsection{Multi-Agent LLM systems' application domain}

The evolution from single-agent to multi-agent LLM systems represents a paradigmatic shift toward distributed intelligence architectures that include collaboration, specialization, and coordination to address complex real-world challenges. Multi-agent frameworks represent a significant evolution in AI capabilities, enabling complex problem-solving through collaboration, specialization, and coordination. This shift has expanded new frontiers in diverse domains where collective intelligence offers advantages in robustness, scalability, and adaptability \cite{chen2023agentverse, ma2024coevolving, zhang2024proagent, huang2025ontheresilience, klein2025fleet}.

\vspace{0.3em} \noindent \textbf{\textit{Healthcare.}} It is among the most impactful domains that benefit from multi-agent LLM-based systems. These systems improve diagnostic accuracy through expert role assignment and collaborative reasoning \cite{kim2024mdagents, chen2025enhancing, ferber2025development} and enhance electronic health record (EHR) processing through coordinated agent interaction \cite{ColaCare_10.1145/3696410.3714877}. Applications also include triage decision support \cite{lu2024triageagent}, scenario-based medical training \cite{barra2025prompt}, and co-dialogue agents for clinical support \cite{chen2025enhancing}. Such systems enable distributed medical reasoning, integrating diverse knowledge modalities while improving interpretability and human-AI alignment.

\vspace{0.3em} \noindent \textbf{\textit{Software engineering and scientific discovery.}} Multi-agent systems are also capable of optimizing complex workflows in the engineering domain. Agents collaboratively design hardware systems \cite{wu2024chateda}, co-develop AI pipelines \cite{xue2025comfybench}, and solve engineering problems through shared knowledge and dynamic role allocation \cite{ni2024mechagents}. In computational biology and chemistry, agents combine domain-specific tools with LLM reasoning for protein discovery \cite{ghafarollahi2024protagents}, molecular analysis \cite{huang2024protchat}, and automated lab planning \cite{m2024augmenting}, highlighting the benefits of combining symbolic, statistical, and simulation-based reasoning between agents. 

\vspace{0.3em} \noindent \textbf{\textit{Social intelligence and cognitive modeling.}} Agentic models offer testing environments to simulate human-like collaboration. Multi-agent frameworks explore cognitive process \cite{li2023camel}, trust dynamics \cite{xie2024can}, prosocial behavior \cite{liu2025exploring}, and social psychology-informed coordination \cite{zhang-etal-2024-exploring}. Several works introduce benchmarking frameworks to assess social intelligence and collaborative behaviors in multi-agent LLM systems \cite{wang-etal-2024-sotopia, chang2024agentboard}. Others focus on reflective collaboration, where agents monitor and revise their strategies \cite{bo2024reflective}, or compete for performance gains \cite{CompeteAI_10.5555/3692070.3694596}. These advances contribute to the development of agents that exhibit nuanced social behaviors and adaptive communication \cite{wang-etal-2024-sotopia, Selfalignment_10.5555/3692070.3693667}.

\vspace{0.3em} \noindent \textbf{\textit{Robotics and embodied AI}.} Multi-agent LLMs coordinate robotic units in both virtual and physical environments. Agents collaboratively plan multi-robot tasks \cite{10802322_SMART-LLM}, generate adaptive driving simulations \cite{wei2024editable}, and train embodied agents using parallel text world simulations \cite{yang2024embodied}. Expanding on these capabilities, LLM-based multi-agent systems like CoELA demonstrate how language interfaces can enable decentralized planning, communication, and cooperation in complex embodied environments \cite{zhang2024building}.

\vspace{0.3em} \noindent \textbf{\textit{General-purpose multi-agent frameworks.}} The frameworks include co-evolutionary agent training \cite{ma2024coevolving}, platforms for the discovery of emergent behavior \cite{chen2023agentverse}, decentralized collaboration via blockchain \cite{jin2024decoagent}, and robust systems designed to tolerate partial agent failure \cite{huang2025ontheresilience}. These systems manage long-context tasks through sequential delegation \cite{zhang2024chain}, self-optimization of training data \cite{zhou2024star}, and proactive collaborative behavior \cite{zhang2024proagent}. Other frameworks address reflective improvement \cite{bo2024reflective}, analytical decision-making \cite{cheng2024sociodojo}, and social self-improvement through simulated interaction \cite{Selfalignment_10.5555/3692070.3693667}, forming the backbone for more domain-specific deployments.

\vspace{0.3em} \noindent \textbf{\textit{Urban planning and telecommunications.}} Domains such as urban planning and communication are among the few unexplored domains that have also adopted a multi-agent LLM system. Coordinated agents manage large-scale network infrastructures \cite{xu2024large} and satellite communication systems using expert mixtures \cite{zhang2024generative}, demonstrating scalability in mission-critical and real-time environments. Moreover, the multi-agent approach improves specialization and fault tolerance \cite{huang2025ontheresilience}, promotes cumulative problem solving \cite{klein2025fleet}, and allows systems that better reflect human collaborative dynamics \cite{zhang-etal-2024-exploring, xie2024can}. As these systems advance in capability and complexity, they are set to transform the way intelligent agents engage in various research domains and real-world applications. 

\section{Reasoning, planning, and memory of LLM agents}
\label{sec:plan_reason}
This section reviews the core cognitive functions: reasoning, planning, and memory, within agents based on LLM. We compare how these capabilities are designed and utilized across single-agent and multi-agent systems, highlighting common strategies and architectural distinctions. An overview of the reasoning, planning, and memory techniques employed in LLM-based agents in single-agent, multi-agent, and widely used categories is summarized in Table S2 of the Supplementary Material.

\subsection{Reasoning in LLM-based agents}

\subsubsection{Application-specific reasoning techniques}

Application-specific reasoning techniques address the limitations of standard approaches in specialized contexts by incorporating domain-specific knowledge, such as scientific logic, API tool usage, or multi-agent interaction dynamics.

\vspace{0.3em} \noindent \textbf{\textit{Single-agent reasoning techniques.}} A range of reasoning strategies has been designed for single-agent settings where an agent acts, plans, and self-assesses autonomously. TOOLLLM \cite{qin2024toolllm} employed depth-first search-based decision trees to navigate over 16,000 real-world APIs, highlighting the strength of search-based reasoning in tool-rich environments. This reasoning technique outperforms ReACT by approximately 81\% in average pass rate. This result suggests that the primary bottleneck in LLM agent reasoning lies less in algorithmic complexity and more in the model’s capacity for coherent long-horizon planning. EASYTOOL \cite{yuan-etal-2025-easytool} advanced task decomposition by breaking high-level goals into modular sub-problems to improve step-wise execution. It significantly enhances LLM performance by providing concise, structured tool instructions. Compared to methods like ReAct and CoT prompting, it achieves higher success rates with fewer errors, demonstrating superior reasoning and tool-utilization capabilities. MATRIX \cite{Selfalignment_10.5555/3692070.3693667} introduced simulation-based self-critique for introspective evaluation. ToolEmu \cite{ruan2024identifying} introduced risk-prompted reasoning, implemented based on the ReAct framework, which allows GPT-4 to assess the risks of tool execution within a sandboxed setup. It allows systematic risk analysis of LLM agents and expands evaluation capability across various tools and testing configurations. Formal frameworks such as Theorem-of-Thought (ToTh) \cite{abdaljalil-etal-2025-theorem} decompose reasoning into abductive, deductive, and inductive subagents coordinated through belief propagation and NLI-based edge scoring. Compared to widely utilized techniques like CoT, CoT-Decoding, and Self-Consistency, ToTh achieved up to 29\% higher accuracy on symbolic tasks and consistently strong performance across complex symbolic and numerical reasoning benchmarks. 

Cognitive realism is further explored in CogMir \cite{liu2025exploring}, which uses social context prompts to emulate cognitive biases, and moral regret mechanisms \cite{tennant2025moral} introduce ethically aware reasoning processes. Domain-specific adaptations include SHAP-based interpretability for semiconductor prediction \cite{zhang2024large}, beneficial hallucination-driven hypothesis testing in software debugging \cite{feldt2023towards}, and perplexity-informed confidence estimation for DNA sequence interpretation in ChatNT \cite{m2024augmenting}. 

Collectively, these single-agent innovations demonstrate how LLMs can reason, evaluate, and adapt in isolation, often tackling complex tasks that demand high degrees of autonomy, safety, and explainability.

\vspace{0.3em} \noindent \textbf{\textit{Multi-Agent Reasoning Techniques.}} Several frameworks demonstrate specialized reasoning innovations tailored to multi-agent settings, showcasing how LLM agents collectively coordinate, communicate, and reason in complex environments. 

For example, Prompt-Structured Strategic Reasoning \cite{li2025llm} was introduced for multi-agent RL, enabling agents to collaborate in cooperative games effectively. AGENTVERSE \cite{chen2023agentverse} supports collaborative reasoning of multiple agents through structured logical representations such as logic grid puzzles and Modular Grounded Symbolic Modules. It addresses issues such as erroneous feedback and incomplete task coverage. In experiments, GPT-4 agent groups consistently achieve stronger performance than single-agent and CoT approaches across various reasoning tasks. Modeling belief states is essential in dynamic and partially observable domains; a Theory-of-Mind (ToM) inspired belief module \cite{li-etal-2023-theoryofmind} was developed to enhance coordination in multi-agent text-based games. 

Similarly, Thread Memory \cite{jin2024large} enables contextual reasoning in urban simulations involving mixed fleets of autonomous and human-driven agents. These innovations increasingly emphasize social, communicative, and contextual reasoning mechanisms that empower LLM agents to function as coherent participants in distributed, multi-agent ecosystems.

\subsubsection{Widely-used reasoning techniques}

Existing reasoning methods, such as CoT, ReAct, Self-Reflection, and Self-Critique, are crucial to enabling LLMs to perform complex cognitive tasks. These techniques support step-by-step problem solving, adaptive decision-making, and collaborative reasoning, contributing to their widespread adoption and effectiveness in improving LLM reasoning capabilities. CoT prompting \cite{bai2025collaboration, li-etal-2023-theoryofmind, yin-etal-2024-agentlumos, zhu2024bootstrapping, zhang2024proagent, ruan2024identifying, abdaljalil-etal-2025-theorem, liu2025exploring, wu2024chateda, zhang2024generative, hong2024cogagent, yang2024llm, zhai2024fine, wang2024omnijarvis, xu2024large, xue2025comfybench, liu2025drbioright} is one of the most prominent approaches. It decomposes complex problems into intermediate logical steps, demonstrating success across the code generation, healthcare, and robotics domains. 

Another notable technique, ReAct \cite{shinn2023reflexion, Understanding_the_Weakness_10.1145/3637528.3671650, feng-etal-2024-large, Offline_10.5555/3692070.3694566, zhang-etal-2024-timearena} integrates reasoning with real-time task-specific actions, allowing agents to adapt dynamically to environmental feedback. This feature is particularly beneficial in interactive and decision-intensive settings. Furthermore, the Self-Reflection and Self-Critique mechanisms \cite{zhang-etal-2024-timearena, zhang-etal-2024-exploring, liu2025exploring, zhang2024building, zhang2024generative, yang2024llm, Describe_10.5555/3666122.3667602, shinn2023reflexion, behari2024decision, zhai2024fine, qu2024recursive, bo2024reflective, ning2024urbankgent, barra2025prompt, m2024augmenting} are robust and widely used for their ability to empower agents to assess their outputs, identify errors, and improve subsequent reasoning through iterative metacognitive processes. Self-consistency methods \cite{kim2024mdagents, RCAgent_10.1145/3627673.3680016} improve reliability by generating multiple reasoning trajectories and selecting the most coherent response. This process, with variants such as Trajectory Level Self-Consistency, further improves the accuracy of planning. 

In addition, collaborative approaches such as Multi-Agent Collaborative Reasoning \cite{zhang-etal-2024-exploring, lu2024triageagent, ghafarollahi2024protagents, chen2025enhancing} allow interactions between multiple agents to debate and refine conclusions. Lastly, ensemble and consensus strategies, including ensemble refinement and temperature-based ensembles \cite{kim2024mdagents}, improve decision making by aggregating multiple model perspectives and combining in-context learning \cite{ma2024coevolving, fu2024autoguide} for better adaptability and knowledge transfer. 

Comparatively, CoT exhibits strong performance in structured, stepwise reasoning but is less effective for interactive or dynamic tasks. Conversely, ReAct augments adaptability through the integration of reasoning and action, although it is associated with increased implementation complexity. Self-reflection facilitates iterative optimization of outputs but remains contingent upon adequate contextual information. Accordingly, the selection of an appropriate reasoning methodology requires rigorous consideration of task attributes, interaction requirements, and computational constraints.

\subsection{Planning in LLM-based agents}
\subsubsection{Application-specific planning techniques}
By integrating customized strategies, agents can better handle complex scenarios that general planning approaches might not optimally solve. We have categorized the existing planning methods into multi-agent and single-agent strategies. 

\vspace{0.3em} \noindent \textbf{\textit{Single-agent planning techniques.}} Single-agent frameworks emphasize self-sufficient decision making and autonomous planning that support independent agent functionality. Examples include TIMEARENA \cite{zhang-etal-2024-timearena}, which employs heuristic planning through temporal task decomposition to simulate realistic and time-sensitive environments. By addressing the critical issue of coordinating overlapping tasks with limited resources, it enables the evaluation and improvement of language agents’ multitasking and temporal reasoning capabilities. ReHAC \cite{feng-etal-2024-large} applies multistep planning grounded in Markov Decision Processes (MDP), allowing agents to navigate human task allocations using formal probabilistic frameworks. Unlike traditional rule-based or prompt-driven planning models, it comprises RL to generalize across different tasks, datasets, and collaboration paradigms, reducing reliance on extensive human input or domain expertise. Cognitive agents using task stack planning \cite{zhu2024bootstrapping} achieve structured goal decomposition, particularly useful in household or task-intensive domains. 

Single-shot planning \cite{cai2025llm} generates a complete plan in a single step, ideal for time-constrained decision-making contexts. The World Knowledge Model \cite{qiao2024agent} introduces guided planning informed by persistent task knowledge, reducing the dependency on blind exploration by encoding task-specific insights. Lastly, RAFA \cite{Reason_for_future_10.5555/3692070.3693331} separates reasoning about future actions from current behavior through multi-step trajectory generation, enabling proactive planning in sequential decision-making tasks.

\vspace{0.3em} \noindent \textbf{\textit{Multi-agent planning techniques.}} Several studies have proposed multi-agent planning strategies to address the complexities of coordination, delegation, and collaboration between distributed agents. For example, ProAgent \cite{zhang2024proagent} enabled modular planning supported by validation mechanisms to facilitate robust collaboration. This approach outperforms traditional learning-based agents, particularly in zero-shot coordination scenarios where such agents typically struggle. Then, in the recommendation system domain, a plan-first execution paradigm \cite{Recommender_AI_10.1145/3731446} was adopted to improve the reliability of action through comprehensive pre-action planning. DeCoAgent \cite{jin2024decoagent} utilized the decomposition of tasks based on prompts, based on a few shots, to enable decentralized agents to collaborate autonomously through smart contracts, offering a scalable framework for coordination. 

Adaptive planning also appears in causal strategy learning systems like \cite{lampinen2023passive}. This implements a two-phase structure, experimentation followed by exploitation, to allow agents to dynamically adjust strategies over time. AGENTBOARD \cite{chang2024agentboard} introduces a progress rate metric to monitor sub-goal completion across agents, thus supporting continuous evaluation in collaborative settings. It gives a clearer insight into agent capabilities compared to conventional success-rate benchmarks. Star Agents \cite{zhou2024star} adopt an evolutionary strategy planning to enhance data optimization through agent evolution. Structured workflows are demonstrated in UrbanKGent \cite{ning2024urbankgent}, where fixed pipeline planning ensures deterministic construction of knowledge graphs in a multi-agent environment. SMART-LLM \cite{10802322_SMART-LLM} tackles multi-robot coordination through coalition formation planning, dynamically forming agent teams to achieve shared objectives. Human-in-the-loop workflow planning in healthcare simulations \cite{barra2025prompt} enables structured scenario designs, combining multiple agents under expert-defined protocols for controlled execution. 

\subsubsection{Widely-used planning techniques}
Planning methods are vital for agents to perform effectively across diverse tasks, with several robust techniques widely used due to their versatility across domains. Multistep planning enables agents to sequence actions over extended horizons, adapting dynamically based on intermediate outcomes and environmental feedback \cite{feng-etal-2024-large, cai2025llm, 10802322_SMART-LLM, Reason_for_future_10.5555/3692070.3693331}. Task decomposition planning breaks down complex goals into manageable subtasks, facilitating efficient resource allocation and parallel execution \cite{wei2024editable, jin2024decoagent, yuan-etal-2025-easytool, zhang-etal-2024-timearena}. Its main advantage is the reduction of overall problem complexity, but performance can degrade when subtasks are highly interdependent, requiring careful sequential coordination to ensure correct execution. ReAct-based planning integrates reasoning with action, allowing agents to adjust plans in real time according to environmental feedback \cite{Offline_10.5555/3692070.3694566, zhang-etal-2024-timearena, singh2024personal}. While it can improve adaptability, its effectiveness varies with task complexity and depends heavily on accurate environment modeling. Reflexion planning incorporates feedback loops and self-correction mechanisms, helping agents learn from errors and improve future performance \cite{zhang-etal-2024-timearena, singh2024personal, feng-etal-2024-large}. Re-planning and dynamic adaptation techniques enable agents to revise strategies responsively when initial plans fail or conditions change unexpectedly \cite{song2023llm, Describe_10.5555/3666122.3667602}, which can improve robustness in uncertain environments, though frequent replanning may introduce computational overhead. Furthermore, plan generation and evaluation approaches improve overall planning quality by creating and assessing multiple candidate plans prior to execution \cite{wang-etal-2024-sotopia, huang2023benchmarking}. 

\subsection{Memory mechanisms in LLM agents}

\subsubsection{Application-specific memory techniques}

\vspace{0.3em} \noindent \textbf{\textit{Single-agent memory techniques.}} Single-agent LLM systems emphasize memory techniques that optimize resource management, contextual awareness, and episodic continuity within individual agent frameworks. For example, RE-Bench, proposed by Wijk et al., tracks background memory during training \cite{wijk2025rebench}, and RCAgent's OBSK manages the structured context \cite{RCAgent_10.1145/3627673.3680016}. Healthcare dialogue benefits from memory buffers \cite{cuadra2024digital}, while VLN-Imagine augments vision transformer memory with generated images \cite{Perincherry_2025_CVPR}. 

Temporal memory tools include prompt context tracking in ToolEmu \cite{ruan2024identifying} and compressed Research Logs in MLAgentBench \cite{huang2023benchmarking}. Formal Reasoning Graphs \cite{abdaljalil-etal-2025-theorem} and STEVE-EYE’s episodic re-prompting \cite{zheng2024steveeye} support transparency and continuity. Specialized memory includes moral alignment prompts \cite{tennant2025moral}, extended histories in AGENTBENCH \cite{liu2024agentbench}, and integration of external screenshots and latent weights in CogAgent \cite{hong2024cogagent}. 

Additional methods involve rolling logs for robots \cite{frering2025integrating}, previous tool output in context \cite{yang2024llm}, negotiated ontologies to prevent repetition \cite{zhang2024large}, and concatenated dialogue contexts for interactive planning \cite{Describe_10.5555/3666122.3667602}. These techniques improve adaptability and long-term task performance in single-agent systems. 

\vspace{0.3em} \noindent \textbf{\textit{Multi-agent memory techniques.}} Application-specific memory techniques in multi-agent LLM systems have been developed to address the complexities of collaborative and distributed environments, employing advanced retrieval and dynamic management strategies tailored to these settings. 

The recent K and relevant K retrieval strategies of ProAgent \cite{zhang2024proagent} and ColaCare's retrieval-augmented generation module \cite{ColaCare_10.1145/3696410.3714877} exemplify efficient memory access. Domain-specialized structures include satellite network knowledge bases segmented into sub-blocks \cite{zhang2024generative}. Dynamic memory approaches involve refeeding prior dialogue turns for attitude shifts \cite{liu2025exploring}, discounted back-tracking for state revival \cite{klein2025fleet}, and running textual memory in MATRIX \cite{Selfalignment_10.5555/3692070.3693667}. Long-term persistent storage for agent states and blockchain collaboration is implemented in DeCoAgent \cite{jin2024decoagent}, while CompeteAI maintains daybook and comment histories to analyze strategies over time \cite{CompeteAI_10.5555/3692070.3694596}. 

These systems emphasize coordinated memory for inter-agent planning and sustained reasoning. Moreover, these approaches demonstrate the critical importance of memory architectures that support inter-agent coordination, long-term strategic planning, and specialized domain reasoning in multi-agent LLM systems.

\subsubsection{Widely-used memory techniques}
Various memory techniques have been proposed to support LLM agents in different operational contexts, each addressing specific information retention and recall requirements. Context window memory is one of the most commonly used methods \cite{zhang2023large, pang2024kalm, wang2024omnijarvis, qu2024recursive, ning2024urbankgent, xie2024can, 10802322_SMART-LLM, song2023llm, Understanding_10.1145/3544548.3581503, boiko2023autonomous, goodell2025large}. Using the internal mechanisms of the transformer, it retains recent inputs and preserves local coherence throughout the prompt. This approach enables agents to maintain consistency during short interactions without relying on external systems. In contrast, conversational and dialogue history stores complete sequences of past interactions to support context-aware responses and maintain continuity over extended sessions \cite{wang-etal-2024-sotopia, li2023camel, Understanding_10.1145/3544548.3581503, wei2024editable, ghafarollahi2024protagents, barra2025prompt, chen2025enhancing}. 

More complex memory implementations include Reflexion-style memory systems, which integrate recent context with structured representations of prior outcomes, allowing agents to update behavior based on previous successes and failures \cite{singh2024personal, feng-etal-2024-large, shinn2023reflexion, bo2024reflective}. Furthermore, RAG methods connect agents to external knowledge sources during inference, enabling access to factual information \cite{ColaCare_10.1145/3696410.3714877, jin2024decoagent, barra2025prompt, goodell2025large}.
The components of the working memory and the scratchpad temporarily hold intermediate reasoning steps and enable sequential problem solving \cite{abdaljalil-etal-2025-theorem, cai2025llm}. In parallel, long-term episodic memory models iteratively capture and organize agent experiences, supporting generalization of past situations and improving task adaptation \cite{zhang2023large, yang2024embodied}. Specific systems apply short-term memory optimization techniques for immediate reactivity tasks to focus on rapid access to updated data while minimizing overhead \cite{bo2024reflective, Understanding_10.1145/3544548.3581503, xu2024large, huang2024protchat}. Several systems implement hybrid memory mechanisms, including models that merge local prompt-based memory with persistent indexed documents \cite{boiko2023autonomous}. These designs reflect ongoing efforts to construct memory architectures that support immediate responsiveness and long-term behavioral consistency.

Different memory mechanisms vary in effectiveness according to the task and the context of the agent. Context window memory suits short-term, reactive tasks, while conversational and episodic memory support extended interactions and continuity. Retrieval-augmented and hybrid systems improve access to external knowledge, enhancing reasoning in complex domains. In multi-agent settings, structured and dynamic memory enable coordinated planning. In general, memory designs are most effective when aligned with specific operational demands.

\section{Impact of prompting, fine-tuning, and memory augmentation}
\label{8_impacts}

Although foundational LLMs have extensive general knowledge and strong reasoning skills, they are not automatically independent agents. Turning these general models into specialized, goal-driven agents that can see, plan, and act well requires a variety of advanced techniques\cite{jin2024decoagent,lu2024triageagent,feng-etal-2024-large}. The performance, reliability, and independence of LLM-based agents are highly dependent on three key areas of improvement: prompting, fine-tuning, and memory enhancement. These methods help developers shape the agent's behavior, give it expertise in specific areas, and prepare it for complex long-term tasks \cite{ghafarollahi2024protagents,huang2024protchat,abbasian2025conversational}.

\subsection{Prompt engineering: a non-parametric approach to dynamic control and role delegation}

Prompt engineering has become a key method for directing the behavior of LLM-based agents \cite{barra2025prompt,li2025eliciting,george-etal-2024-probing}. Unlike parametric approaches like fine-tuning, which change a model's weights, prompt engineering works during inference time. It offers a simple way to define tasks, assign roles, and control agent behavior dynamically without needing a lot of computing power. This method takes advantage of the natural instruction-following and in-context learning abilities of LLMs to manage complex tasks \cite{abdaljalil-etal-2025-theorem}. This includes everything from individual agent reasoning to complex teamwork among multiple agents \cite{kim2024mdagents}. The flexibility of prompting makes it the primary way to set goals, integrate tools, and simulate complex social interactions. At the single-agent level, prompt engineering plays a key role in defining an agent's operational settings and reasoning style. For instance, Reflexion \cite{shinn2023reflexion} demonstrates how prompts enable self-reflection without weight updates, achieving 
performance improvements through feedback-guided adjustments. This type of self-correction, guided by prompts, also works well in dynamic situations. Behari et al. \cite{behari2024decision} used a Decision-Language Model (DLM) to understand natural language policy goals, suggest reward functions as code, and adjust them based on the results of simulations. However, the effectiveness of this control can be limited by context length and complexity. George et al. \cite{george-etal-2024-probing} notes that agents may struggle to use important information when faced with long histories and distracting "red herring" facts, even with CoT prompting. Moreover, prompting is the main way to enable agents to use external tools, which enhances their abilities beyond their existing knowledge. EASYTOOL \cite{yuan-etal-2025-easytool} converts extensive tool documentation into clear, simple instructions. This helps LLMs choose and perform the correct functions and has been successfully applied in specialized fields. Besides, ChemCrow \cite{m2024augmenting} combines 18 expert-designed chemistry tools that an LLM agent learns to use through well-designed prompts to independently plan and carry out complex syntheses. In the medical field, Goodell et al. \cite{goodell2025large} show that giving agents task-specific calculation tools like OpenMedCalc via prompting greatly reduces math errors. These domain-specific designs, ChemCrow and OpenMedCalc, yield higher precision in their respective fields yet require substantial expert knowledge. Therefore, selecting a prompt-tool strategy depends on whether adaptability or domain performance is prioritized. This concept extends to multimodal contexts, where Yang et al. \cite{yang2024llm} apply prompts in LLM-Grounder to break down language queries and manage visual grounding tools. Meanwhile, Song et al. \cite{song2023llm} connect agent plans in LLM-Planner by updating prompts with lists of visually perceived objects. Prompt engineering ranges from controlling individual agents to managing entire "societies" of agents. Here, system-level prompts set roles, responsibilities, and communication protocols. Ni et al. \cite{ni2024mechagents} show this with the MechAgents framework, which creates teams of agents with roles such as planner, coder, and critic to solve mechanics problems. Chen et al. \cite{chen2023agentverse} propose AGENTVERSE to create expert roles and task instructions in a zero-shot manner. To manage long-context tasks, Zhang et al. \cite{zhang2024chain} introduce the Chain-of-Agents (CoA) framework, where a manager agent combines contributions from multiple worker agents, each handling a part of the document. Likewise, Klein et al. \cite{klein2025fleet} suggest the FLEET OF AGENTS (FOA) framework, which uses a particle filtering method where many agents explore a search space, guided by prompts to improve the cost-quality balance. 
This approach has proven very effective in simulating complex social and economic systems that resemble human behavior. Zhang et al. \cite{zhang-etal-2024-exploring} build LLM agent societies with distinct traits and thinking patterns using only prompts. These societies display behaviors such as conformity and consensus. Zhao et al. \cite{CompeteAI_10.5555/3692070.3694596} employ prompts in CompeteAI to create a virtual town filled with competing restaurant agents to examine market dynamics. Liu et al. \cite{liu2025exploring} present the CogMir framework, which uses prompts to encode social variables and induce prosocial irrationality, showing that LLM agents can repeatedly mimic human cognitive biases. This multi-agent, prompt-driven method has strong applications in specific fields. In clinical triage, Lu et al. \cite{lu2024triageagent} use TRIAGEAGENT to simulate a multi-disciplinary team. Chen et al. \cite{chen2025enhancing} apply a similar method, MAC, for diagnosing rare diseases. In scientific discovery, Ghafarollahi et al. \cite{ghafarollahi2024protagents} organize agents in ProtAgents to design new proteins together. Barra et al. \cite{barra2025prompt} create a workflow to automate the production of healthcare simulation scenarios. While prompt engineering is a strong non-parametric tool, it also works well with parametric fine-tuning. Prompting often serves as a way to create high-quality, structured datasets for training smaller, specialized agents. For example, Qin et al. \cite{qin2024toolllm} use ChatGPT in their ToolLLM framework to generate a large dataset of instructions and API call solutions. This dataset is then used to fine-tune LLaMA into the effective ToolLLaMA. Similarly, Qian et al. \cite{qian-etal-2024-tellmemore} develop Mistral-Interact by fine-tuning Mistral-7B on a set of user dialogues created by prompting GPT-4. Pang et al. \cite{Selfalignment_10.5555/3692070.3693667} use the MATRIX framework to generate alignment data for fine-tuning. This method, as Wu et al. \cite{wu2024chateda} point out, allows the skills of large, proprietary models to be distilled into smaller, open-source agents. On the other hand, fine-tuning can help agents better follow complex prompts or perform specific skills. Yin et al. \cite{yin-etal-2024-agentlumos} demonstrate that fine-tuning agents with the LUMOS framework on unified, high-quality annotations improves their planning and grounding abilities. Wang et al. \cite{wang-etal-2024-sotopia} fine-tune a 7B model in SOTOPIA-\(\pi\) using data from prompted social interactions. This makes it match the social intelligence of a much larger GPT-4-based agent. This technique proves particularly useful in embodied AI. Yang et al. \cite{yang2024embodied} show that an LLM expert, enhanced through reflective prompting, can generate distillation data to fine-tune a Vision-Language Model (VLM), teaching it to navigate and interact with a visual world. This shows that while prompting offers dynamic, real-time control, fine-tuning can embed more robust and specialized abilities into agents.

\subsection{Fine-tuning: embedding domain expertise and core behavioral traits}

While prompt engineering allows for quick control during inference, fine-tuning offers an additional method to embed specialized knowledge and important traits into an agent's design \cite{kim2024mdagents,zheng2024steveeye,klein2025fleet}. This process updates the model's weights using curated datasets. It is essential for developing smaller, more efficient agents that can imitate the abilities of larger models or for teaching complex skills that are difficult to achieve through prompting alone \cite{lu2024triageagent}. Fine-tuning is the main way to adapt to new fields, gain skills, and improve the basic reasoning and interaction abilities of agents based on LLMs \cite{zhang2024building,wu2024chateda,zhang2023large}.
A common practice in developing agents is using fine-tuning for knowledge distillation. In this method, the advanced reasoning of a large, skilled model generates high-quality training data for a smaller, open-source model \cite{lu2024triageagent}. This makes powerful capabilities more accessible. For example, Qin et al. \cite{qin2024toolllm} pioneered this with ToolLLaMA by fine-tuning LLaMA on a vast dataset of tool-use instructions and solutions from ChatGPT. This led to a highly capable open-source tool-using agent. Qian et al. \cite{qian-etal-2024-tellmemore} created Mistral-Interact by fine-tuning Mistral-7B on user dialogues simulated by GPT-4. It focuses on enhancing user-agent interaction, proactively clarifying vague instructions, and refining user intentions before task execution. Pang et al. \cite{Selfalignment_10.5555/3692070.3693667} applied the MATRIX framework to produce social simulation data for alignment fine-tuning. This method also works for specialized fields, as shown by Wu et al. \cite{wu2024chateda} with ChatEDA, where a fine-tuned model, AutoMage, outperformed GPT-4 in electronic design automation tasks. The process is often iterative. Frameworks like Star-Agents from Zhou et al. \cite{zhou2024star} use multi-agent systems to automatically generate, evaluate, and refine instruction data to improve the fine-tuning process itself. Beyond simple knowledge transfer, fine-tuning is essential for developing the complex behavioral and cognitive skills that define an agent's main abilities. This includes social intelligence, where Wang et al. \cite{wang-etal-2024-sotopia} used data from guided social interactions to fine-tune SOTOPIA-\(\pi\). This approach enabled a 7B model to achieve social goal completion similar to that of a larger GPT-4 agent. Fine-tuning can also improve self-awareness skills. Qu et al.\cite{qu2024recursive} created RISE, an iterative fine-tuning process that helps agents to think about their own actions and fix their mistakes, which is difficult to accomplish through prompting alone. Similarly, Bo et al.\cite{bo2024reflective} fine-tuned a shared "reflector" agent within their COPPER framework to enhance collaboration among multiple agents. This method works particularly well for embodied agents, as fine-tuning connects abstract reasoning with physical action. Zhai et al.\cite{zhai2024fine} used RL to fine-tune VLMs, improving their decision-making in goal-oriented tasks. Yang et al.\cite{yang2024embodied} employed an LLM expert to create distillation data to fine-tune a VLM called EMMA, teaching it to understand and interact with a visual environment. Fine-tuning is also crucial for transforming general models into specialized experts that can function in complex, real-world settings. In the realm of embodied AI, Schumann et al.\cite{schumann2024velma} demonstrated that while prompting VELMA had some success, fine-tuning on navigation examples resulted in a 25\% relative improvement in task completion. Hong et al.\cite{hong2024cogagent} fine-tuned CogAgent on GUI-grounding data, greatly enhancing its ability to navigate and manage computer interfaces. This specialization is vital in the fields of science and medicine. Liu et al.\cite{liu2025drbioright} fine-tuned DrBioRight 2.0 on a large cancer proteomics dataset to develop an expert bioinformatics chatbot. In addition, de Almeida et al.\cite{de2025multimodal} fine-tuned ChatNT to create a multimodal agent with a deep understanding of DNA, RNA, and protein sequences. However, embedding expertise comes with risks; Yang et al.\cite{yang2024watch} showed that the fine-tuning process itself can be an avenue for backdoor attacks, where tainted data might train an agent to carry out harmful actions secretly. Ultimately, fine-tuning and prompting work together in a complementary way. Good prompting often serves as a foundation for creating the high-quality data needed for effective fine-tuning. On the other hand, fine-tuning can help agents respond better to complex prompts and improve their planning. Yin et al.\cite{yin-etal-2024-agentlumos} illustrate this with the LUMOS framework, demonstrating that fine-tuning on unified, high-quality annotations enhances both planning and grounding skills. This interaction facilitates the development of strong, specialized, and efficient agents where parametric training builds basic skills and non-parametric prompting guides their active use during inference.

Moreover, fine-tuning introduces risks, including model overfitting, amplification of biases present in the training data, and susceptibility to malicious data injection. While fine-tuning can achieve higher task-specific performance, it requires careful dataset curation, extensive computational resources, and rigorous validation. In contrast, prompt engineering offers flexibility and safety, but generally cannot reach the same depth of expertise.

\subsection{Memory augmentation: enabling grounded reasoning and experiential learning}  

While prompting gives immediate instructions for an agent, memory augmentation extends agents beyond context limits\cite{lu2024triageagent,jin2024decoagent} by changing them from simple instruction-followers into adaptable systems\cite{jin2024large}. Memory can be divided into two main types: retrieval of external knowledge (RAG) and accumulation of internal dynamic experience that supports learning and self-correction. RAG grounds agent reasoning in verifiable information. It helps reduce mistakes and improve reliability. For example, in clinical settings, frameworks like TRIAGEAGENT\cite{lu2024triageagent} and ColaCare\cite{ColaCare_10.1145/3696410.3714877} use RAG to give agents access to medical handbooks and guidelines. This ensures their decisions are based on solid evidence. This method works well in specialized technical areas; Barra\cite{barra2025prompt} employs RAG to offer established simulation guidelines for scenario design. Xia et al.\cite{xia2024generation} use it to access technical datasheets for creating digital twin models. However, the effectiveness of RAG can depend on the model. Xia et al. mention a “cheat-sheet effect,” where it greatly improves performance in weaker LLMs compared to stronger ones. This suggests that the value of external memory depends on the agent’s inherent reasoning capabilities. The external memory can also be dynamic, as shown in frameworks like DeCoAgent\cite{jin2024decoagent}. Here, a JSON memory module pulls current on-chain data, preventing unnecessary blockchain scans. Beyond retrieving static facts, memory is vital for experiential and reflective learning. One important framework in this area is Reflexion\cite{shinn2023reflexion} that stores verbal reflections in episodic buffers. This helps guide future attempts. This process of self-correction has been shown to be effective in many applications, from improving public health policies in simulations\cite{behari2024decision} to generating high-quality distillation data for training other models\cite{yang2024embodied}. However, compared to an episodic system like Reflexion, RAG is more reliable for factual grounding but less effective for long-term adaptation. The concept has evolved into more organized, long-lasting memory systems. REMEMBERER\cite{zhang2023large} updates an agent's long-term experience memory through RL. AVATAR \cite{wu2024avatar} maintains a "Memory Bank" to store high-performing action sequences and instructions. Similarly, Star-Agents\cite{zhou2024star} uses an "Instruction Memory Bank" to continuously improve data generation strategies. Memory also helps agents stay grounded in their immediate, dynamic environment, especially in interactive settings. This is often done by updating the agent's prompt with real-time perceptual information. For example, Song et al.\cite{song2023llm} keep plans grounded in LLM-Planner by constantly updating the prompt with a list of visually perceived objects. Yang et al.\cite{yang2024llm} use prompts to coordinate visual grounding tools. This grounding includes more abstract representations of the state as well. Li et al.\cite{li-etal-2023-theoryofmind} show that giving agents an explicit belief state representation greatly enhances planning and reduces errors in Theory of Mind tasks. In multi-agent systems, memory can be distributed. The CoA framework from Zhang et al.\cite{zhang2024chain} passes a "Communication Unit" in sequence between agents, allowing each to build on the previous work. A more complex version is CoELA\cite{zhang2024building}, which resembles human thinking with distinct semantic, episodic, and procedural memory modules to support long-term cooperation. Despite its advantages, memory enhancement is not perfect and faces significant challenges. The main limitation lies in how well the agent can manage long and complex memory streams. George et al.\cite{george-etal-2024-probing} points out that even with CoT prompting, agents can struggle to use crucial information from lengthy histories filled with distracting facts. Additionally, some memory strategies can backfire in specific situations. Zhang et al.\cite{zhang-etal-2024-timearena} discovered that a Reflexion-style memory system led to worse performance in complex multitasking scenarios. Sometimes, simpler approaches like a "sliding window" memory can be a more effective, if less advanced, solution\cite{chang2024agentboard}. Ultimately, how well memory is implemented is what distinguishes an agent as a reactive tool versus a cognitive entity capable of learning, adapting, and maintaining coherent long-term behavior.
\vspace{-10pt}

\subsection{The synergistic integration of prompting, fine-tuning, and memory}

The development of capable LLM-based agents relies on the interaction of prompting, fine-tuning, and memory augmentation\cite{ni2024mechagents,chen2023agentverse,zhang-etal-2024-timearena}. The combined use of these three elements is crucial for increasing agent autonomy, from single-agent reasoning to complex multi-agent collaboration\cite{kim2024mdagents,ni2024mechagents,ghafarollahi2024protagents}. At its core, prompt engineering is the main method for guiding agent behavior. However, relying solely on prompting can be fragile. Performance drops significantly when agents deal with long contexts filled with distracting "red herring" facts, even with CoT prompting. While prompting provides dynamic control, fine-tuning offers a method for permanently embedding specialized knowledge and complex skills in an agent. This is essential for creating agents that are trained by instructions and are inherently capable in specific domains. One of the most effective combinations occurs when prompting serves as a tool for generating data, creating high-quality datasets for refining smaller, more efficient agents. This model allows the reasoning abilities of large proprietary models to be distilled into open-source alternatives. A clear example is the ToolLLM framework \cite{qin2024toolllm}, which uses ChatGPT to produce a vast dataset of instructions and API call solutions to fine-tune LLaMA into the highly effective ToolLLaMA. A similar method is used by Qian et al.\cite{qian-etal-2024-tellmemore} to develop Mistral-Interact by fine-tuning on user dialogues generated by prompting GPT-4. Pang et al.\cite{Selfalignment_10.5555/3692070.3693667} also use the MATRIX framework to create alignment data. This bidirectional relationship extends further: fine-tuning can enhance an agent's ability to follow complex prompts and execute domain-specific instructions more reliably. Yin et al.\cite{yin-etal-2024-agentlumos} demonstrate this with the LUMOS framework, showing that fine-tuning on unified, high-quality annotations improves both planning and grounding abilities when combined with structured prompting strategies. 

Memory augmentation bridges prompting and fine-tuning mechanisms by providing the contextual foundation necessary for effective prompting and the experiential data required for targeted fine-tuning. RAG-based systems like TRIAGEAGENT\cite{lu2024triageagent} and ColaCare\cite{ColaCare_10.1145/3696410.3714877} use external memory to ground prompts in factual medical knowledge, mitigating hallucinations during inference. Conversely, episodic memory systems like Reflexion \cite{shinn2023reflexion} accumulate task-specific experience in the form of episodic verbal reflections, which are used in subsequent prompts to improve both immediate reasoning and performance. The most advanced agents successfully combine all three components. In these systems, fine-tuning develops core skills, prompting guides inference-time reasoning and tool use, while memory offers dynamic context and knowledge. The CoELA framework\cite{zhang2024building} fine-tunes CoLLAMA using data collected by agents and applies structured prompts alongside a multi-part memory system (semantic, episodic, procedural) to enable complex collaboration. OmniJARVIS\cite{wang2024omnijarvis} demonstrates this integration by fine-tuning a VLA model on unified tokens that represent instructions, memories, and actions. These tokens are coordinated by prompts for open-world tasks. These systems illustrate that, while each mechanism is effective independently, their combination is essential for developing truly autonomous, capable, and reliable LLM-based agents.

\section{Evaluation benchmarks and datasets}
\label{9_eval}
The rapid growth of LLM agents has required a change in how we evaluate them. Early methods that depended on static NLP benchmarks do not effectively capture the interactive, goal-driven, and often multi-step nature of these systems. The literature shows a clear shift towards more dynamic, specific, and analytical evaluation methods\cite{zhang-etal-2024-timearena,wijk2025rebench,wang-etal-2024-sotopia}.

To give a clear overview of this progress, this section looks at the evaluation landscape through three connected lenses. First, it reviews the specialized benchmarks and interactive environments designed to test the abilities of agents in realistic situations\cite{chang2024agentboard,lu2024triageagent}. Next, it examines the various methods and metrics developed to judge performance beyond simple accuracy, including factors like efficiency, quality, and behavioral strength\cite{chang2024agentboard}. Finally, the section highlights the important role of the datasets used to train and ground these agents since they provide the basis for agentic behavior.

\subsection{Task-oriented and interactive benchmarks}

Evaluating the capabilities of LLM-based agents requires a shift from static language metrics to dynamic benchmarks that focus on task performance and interactivity in complex environments. A notable trend is the creation of simulated environments where agents must engage in multi-step reasoning and actions. For example, TIME-ARENA\cite{zhang-etal-2024-timearena} introduced a text-based simulation that includes time-based dynamics and constraints, challenging agents with multitasking scenarios in cooking, household tasks, and lab work. Advancing this complexity, AndroidArena\cite{Understanding_the_Weakness_10.1145/3637528.3671650} offers a general-purpose operating system environment to assess agents on intricate tasks that need cooperation between applications and compliance with user constraints. In a more specific area, RE-Bench provides demanding, open-ended machine learning research tasks, allowing for a direct comparison between AI agents and human experts. Environments like Overcooked-AI, ALFRED, and WebShop are established settings for testing zero-shot coordination, instruction following, and web navigation, respectively \cite{zhang2024proagent, song2023llm,klein2025fleet}.

An important part of agent functionality is their ability to interact with and manipulate external tools and APIs. ToolLLM introduced ToolBench, which is a detailed instruction-tuning dataset with over 16,000 real-world APIs, along with an automatic evaluator called ToolEval to assess pass rates and solution quality\cite{qin2024toolllm}. Frameworks like EASYTOOL aim to improve agent performance on these benchmarks by converting extensive tool documentation into brief, clear instructions \cite{yuan-etal-2025-easytool}. The assessment of tool use also includes specialized scientific fields. For instance, ChemCrow and Coscientist are agent systems evaluated on their ability to plan and carry out complex chemical syntheses using expert-designed chemistry tools\cite{m2024augmenting,boiko2023autonomous}. In industrial applications, RCAgent showcases a tool-enhanced agent for cloud root cause analysis, evaluated using proprietary system log data\cite{RCAgent_10.1145/3627673.3680016}.

Beyond executing tasks, researchers are creating benchmarks to evaluate more subtle human-like and social behaviors. The IN3 benchmark focuses on assessing an agent’s capacity to understand user intentions by prompting them to ask clarifying questions\cite{qian-etal-2024-tellmemore}. To examine the limits of agent reasoning, the OEDD corpus presents scenarios where agents must make sense of various experiences while ignoring misleading information\cite{george-etal-2024-probing}. Assessing collaborative and social intelligence is another critical area. SOTOPIA-\(\pi\) features an interactive learning approach along with an evaluation suite, SOTOPIA-EVAL, which employs both human and LLM-based assessments on aspects like goal completion and believability\cite{wang-etal-2024-sotopia}. Additionally, frameworks have been developed to create “societies” of LLM agents to observe collaborative behaviors on tasks from the MMLU, MATH, and Chess datasets. The ability for agents to mimic human behavior is tested in frameworks that use Trust Games to compare agent decisions with recognized human patterns.

Several meta-evaluation frameworks have been created to consolidate and analyze agent performance across various tasks. AgentBoard\cite{chang2024agentboard} offers an evaluation board that includes nine different multi-turn, partially observable tasks from areas like embodied AI, gaming, and web navigation. It provides a detailed metric to track progress, offering deeper insights beyond basic success rates. In the medical field, new benchmarks are emerging to ensure clinical safety and effectiveness. TRIAGE AGENT\cite{lu2024triageagent} released the first public benchmark for clinical triage, using metrics like discordance and undertriage rates against human expert performance. Other studies assess agents in their ability to execute clinical calculations using established medical calculators or make complex oncology decisions based on multimodal patient data\cite{goodell2025large,ferber2025development}. These interactive benchmarks are essential for fostering the growth of more capable, reliable, and aligned LLM agents.

\subsection{Methodologies and metrics for evaluation}
The evaluation of LLM-based agents is quickly changing from basic accuracy checks to more complex methods that look at task performance, reasoning quality, and interaction dynamics\cite{wang-etal-2024-sotopia,li-etal-2023-theoryofmind,chang2024agentboard}. A key approach continues to focus on task-oriented performance metrics, where we measure an agent’s success by its ability to reach specific goals\cite{lu2024triageagent,qin2024toolllm}.

This is often assessed through accuracy or success rates on established benchmarks for coding, like HumanEval, math reasoning datasets such as GSM8K and MATH, and embodied AI tasks in settings such as ALFRED and Overcooked-AI\cite{chen2023agentverse,ma2024coevolving,song2023llm,zhang2024proagent}. In addition to these general benchmarks, evaluations are frequently customized for specific fields, using metrics like discordance and undertriage rates in clinical triage, the emergence of macroeconomic laws in economic simulations, and improvements in material properties for scientific discovery\cite{goodell2025large,ferber2025development,li-etal-2024-econagent}.

For tasks where success is more subjective or quality has shades of meaning, researchers are turning to human experts and "LLM-as-a-Judge" frameworks. This approach is essential for evaluating things like the quality of travel plans, the realism and empathy of synthetic medical dialogues, and the solution quality for complex tool-use instructions. Metrics in this area often involve comparative judgments, such as win/tie/lose rates against baseline models or preference rates where evaluators choose the better of two outputs. Frameworks like SOTOPIA-EVAL use both human and LLM-based judgments across various dimensions, including believability and relationship development, to evaluate social intelligence\cite{wang-etal-2024-sotopia}.

Evaluating multi-agent systems brings in metrics that capture the complexities of teamwork, social dynamics, and system-level properties\cite{ni2024mechagents,wang-etal-2024-sotopia}. Performance is often measured by team scores, completion times, and comparisons between collaborative groups and solo agents\cite{zhang-etal-2024-timearena,chen2023agentverse}. Methods also explore emerging social behaviors, such as conformity and reaching consensus in agent societies, or the ability to simulate human trust behaviors in economic games\cite{zhang-etal-2024-exploring,li-etal-2024-econagent}. The resilience of these systems is another major concern, with some evaluations measuring how performance degrades when faulty or malicious agents join the collaborative process\cite{huang2025ontheresilience}.

In addition to these outcomes, a growing number of studies focus on the efficiency and internal reasoning processes of agents. Metrics like token consumption, tool invocation efficiency, and the balance between cost and quality are used to evaluate how resourceful agent solutions are \cite{yuan-etal-2025-easytool,klein2025fleet}. At the same time, new methods are being developed to investigate the quality of an agent's reasoning. This includes detailed detection of different types of errors in math problems, assessing ToM accuracy in cooperative games, and evaluating the logical coherence of reasoning sequences using formal graphs\cite{li-etal-2023-theoryofmind,Understanding_10.1145/3544548.3581503}. This attention to internal processes is vital for creating more reliable and interpretable agents.

To create a standard evaluation across different tasks, several meta-evaluation frameworks have been proposed. AgentBoard\cite{chang2024agentboard} provides a comprehensive evaluation platform that combines nine unique multiturn tasks and introduces a detailed progress rate metric for deeper insight beyond simple success rates. Likewise, specialized evaluators like ToolEval\cite{qin2024toolllm} have been created to specifically assess tool-use capabilities, measuring metrics like pass rates and solution quality across thousands of real-world APIs. These initiatives indicate a maturation in the field, moving toward more comprehensive and insightful assessment tools that can help develop more capable and aligned agents.

\subsection{Datasets for agent training and grounding}
The development of effective LLM agents heavily depends on the availability of diverse, high-quality datasets for training and grounding in specific areas. These datasets are rapidly changing, moving away from traditional static text collections to dynamic, interactive, and domain-specific resources. This shift helps agents acquire complex skills and emphasizes grounding their abilities in real-world tasks, tools, and social contexts. Table~\ref{tab:datasets} presents an overview of sixty-eight publicly available datasets employed to train, evaluate, or benchmark LLMs in agent modeling, organized by task type and citation.

\begin{table*}[ht!]
\centering
\caption{Details of the sixty-eight publicly available datasets analyzed in this study, including their limitations and type.}
\label{tab:datasets}
\begin{scriptsize}
\setlength{\tabcolsep}{4pt} 
\begin{tabular}{@{} p{3cm} p{3cm} p{2.8cm} | p{3cm} p{3cm} p{2.8cm} @{}}
\toprule
\textbf{Dataset} & \textbf{Dataset Type} & \textbf{Limitations} &
\textbf{Dataset} & \textbf{Dataset Type} & \textbf{Limitations} \\
\midrule

HumanEval \cite{HumanEval} & Code Generation & Limited Task Diversity &
MBPP \cite{MBPP} & Code Generation & Limited Concept Coverage\\
LeetCodeDataset \cite{LeetCodeDataset} & Code Generation & Narrow Problem Coverage &
CoNaLa \cite{CoNaLa} & Code Generation & Single Answer Bias \\
RepoBench-P \cite{RepoBench} & Code Completion & Limited Language Diversity &
Evol-CodeAlpaca \cite{Evol} & Code Instruction & Limited Domain Coverage \\
InterCode \cite{InterCode} & Interactive Coding & Limited Language Coverage &
GSM8K \cite{GSM8K} & Math Reasoning & Limited Problem Variety \\
MGSM \cite{MGSM} & Math Reasoning & Limited Problemset Coverage &
Math-Shepherd \cite{Math-Shepherd} & Math Reasoning & Narrow Problem Scope \\
MultiArith \cite{MultiArith_problem} & Math Reasoning & Limited Problem Complexity &
MATH \cite{MATH} & Math Reasoning & Lacks Algorithmic Reasoning Problems \\
TabMWP \cite{TabMWP} & Math Word Problem & Limited Problem Diversity &
MoST \cite{MoST} & Multi-Step Reasoning & Inherent Annotation Noise \\
STaRK \cite{STaRK} & Reasoning & Limited Human-Query Variety &
BridgeData \cite{BridgeData} & Reasoning & Limited Task Diversity\\
Aci-bench \cite{Aci-bench} & Agent Interaction & Synthetic Data Reliance &

HotpotQA \cite{HotpotQA} & Multi-hop QA & Limited Domain Coverage \\
MuSiQue \cite{MuSiQue} & Multi-hop QA & Restricted Question Diversity &
StrategyQA \cite{StrategyQA} & Multi-step QA & Limited Factual Diversity \\
ArxivQA \cite{ArxivQA} & Scientific QA & Limited Disciplinary Diversity &
QASPER \cite{QASPER} & Scientific QA & Limited Domain Coverage \\
NarrativeQA \cite{NarrativeQA} & Long-form QA & Limited Training Data &
FEVER \cite{FEVER} & Fact Verification & Limited Evidence Scope \\

QMSum \cite{QMSum} & Query Summarization & Limited Domain Coverage &
WikiHow \cite{WikiHow} & Summarization & Limited Abstraction \\
GovReport \cite{GovReport} & Summarization & Limited Domain Representation &
BookSum \cite{BookSum} & Long Summarization & Limited Coverage Of Diverse Genres  \\

Stanford Alpaca \cite{alpaca} & Instruction Tuning & Limited Contextual Variation. &
OpenAssistant \cite{OpenAssistant} & Instruction Tuning & Limited Domain Coverage \\
OpenOrca \cite{OpenOrca} & Instruction Tuning & Limited Domain Coverage &
WildChat \cite{WildChat} & Instruction Tuning & Limited Real-World Diversity. \\
IN3 \cite{qian-etal-2024-tellmemore} & Instruction Tuning & Limited Task Diversity &

FineWeb \cite{FineWeb} & Pre-training Corpus & Limited Content Diversity \\
RefinedWeb \cite{RefinedWeb} & Pre-training Corpus & Contains Residual Noise  &
RedPajama \cite{RedPajama} & Web Text & Benchmark Contamination Risk  \\

LLMSecEval \cite{LLMSecEval} & Code Security & Limited CWE Coverage 
 &

PKU-SafeRLHF \cite{PKU-SafeRLHF} & Safety & Limited Scale And Harm Granularity \\
HarmfulQA \cite{HarmfulQA} & Safety & Synthetic Prompt Bias
 &
HH-RLHF \cite{HH-RLHF} & Alignment & Limited Generalization Scope \\
BeaverTails \cite{BeaverTails} & Alignment & Narrow Harm Category &
LLM Attacks \cite{LLM_Attacks} & Adversarial & Limited Attack Diversity \\

R2R \cite{r2r} & Embodied AI & Limited Context Diversity &
REVERIE \cite{REVERIE} & Embodied AI & Sparse Object Annotations\\
Touchdown \cite{Touchdown} & Embodied AI & Narrow Urban Context &
ALFRED \cite{ALFRED} & Instruction & Limited Environment Diversity \\
ALFWorld \cite{ALFWorld} & Instruction & Limited Task Variety &
TravelPlanner \cite{TravelPlanner} & Planning & Low Real-World Complexity. \\
BlocksWorld \cite{BlocksWorld} & Planning & Lacks Real-World Variability &

HaGRID \cite{HaGRID} & Gesture Recognition & Lacks Dynamic Gestures \\
EgoGesture \cite{EgoGesture} & Gesture Recognition & Limited Gesture Diversity &

CogScene \cite{CogScene} & 3D Scene Understanding & Lacks Scenario Diversity \\
WIDER FACE \cite{WIDER} & Face Detection & Demographic Bias &
CelebA \cite{CelebA} & Face Attributes &  Limited Attribute Diversity \\

CDSL \cite{CDSL} & Clinical Data & Lacks Temporal Coverage &
CCLE \cite{CCLE} & Genomics & Limited Tissue Representation \\
APARENT2 \cite{de2025multimodal} & Genomics &  Limited Task Complexity &
Saluki \cite{Saluki} & Genomics & Limited Domain Coverage \\
SKEMPI \cite{SKEMPI} & Protein Binding Affinity & Lacks Structural Diversity &
PubMed \cite{PubMed} & Scientific Abstracts & Possesses Inconsistent Indexing \\
GAIA \cite{GAIA} & General AI Assistant & Lacks Language Diversity &

BIG-bench \cite{BIG-bench} & General Evaluation & Limited Task Diversity \\
MMLU \cite{MMLU} & General Evaluation & Data Contains Cultural Bias &
CAMEL \cite{li2023camel} & Multi-Agent Convo. & Limited Temporal Coverage \\
SOTOPIA-PI \cite{wang-etal-2024-sotopia} & Social Simulation & Limited Safety Dimensions &
ToolBench \cite{ToolBench} & Tool Use & Exhibits Tool Specific Bias \\
Mind2Web \cite{Mind2Web} & Web Agent & Evaluation Bias And Limited Scope &

RoboNet \cite{RoboNet} & Trajectory & Limited Task Modalities \\

\bottomrule
\end{tabular}%
\end{scriptsize}
\vspace{-15pt}
\end{table*}


A key strategy involves using and adapting existing public benchmarks to build core reasoning and instruction-following skills. Datasets like those for mathematics (GSM8K, MATH), question-answering (HotPotQA, StrategyQA), and coding (HumanEval) are commonly used for fine-tuning models and serve as a baseline for assessing logical and problem-solving abilities\cite{shinn2023reflexion,yin-etal-2024-agentlumos,huang2025ontheresilience}. Instruction-tuning datasets, such as Alpaca, are used to improve the overall capabilities of models before tailoring them for specific agent tasks\cite{zhou2024star}. To promote broader agent learning, there are ongoing efforts to collect large-scale, unified annotations based on various reasoning frameworks across these complex interactive tasks. A notable trend involves employing agents to automate the enhancement and diversification of these instructional datasets, creating a constant cycle to improve data quality.

A major development in agent training is creating datasets and simulated environments designed for tool use and interaction in complex digital spaces. ToolBench provides an extensive instruction-tuning dataset with over 16,000 real-world APIs, allowing models to learn to execute advanced instructions and adapt to new tools \cite{yang2024watch}. For grounding in interactive settings, benchmarks like WebShop and ALFRED serve as established bases for training and testing web navigation and embodied instruction-following, respectively \cite{klein2025fleet,song2023llm}. More intricate environments, such as AndroidArena \cite{Understanding_the_Weakness_10.1145/3637528.3671650}, assess agents on complex tasks that require cooperation between applications, while TIME-ARENA \cite{zhang-etal-2024-timearena} introduces time dynamics and multitasking challenges in simulated household and lab scenarios. AgentBoard \cite{chang2024agentboard} combines nine distinct multi-turn, partially observable tasks from areas like embodied AI, gaming, and web navigation into one evaluation framework.

To ground agents in specialized, high-stakes fields, researchers are curating and generating domain-specific datasets. In the medical sector, this includes datasets derived from clinical triage manuals (TRIAGE AGENT), public EHR data such as MIMIC-IV, patient engagement logs, and extensive cancer proteomics resources such as TCGA and CCLE \cite{lu2024triageagent,ColaCare_10.1145/3696410.3714877,liu2025drbioright}. In science and engineering, agents are trained on tailored datasets drawn from scientific literature for research in organic semiconductors and chemistry or data generated from physics-based simulations for mechanics problems\cite{boiko2023autonomous,ni2024mechagents}. Similarly, specialized collections are being built for urban planning using public data from cities like San Francisco, for finance using public datasets like MovieLens and Amazon-Beauty to simulate user behavior, and for electronics design automation by generating custom code-instruction pairs\cite{jin2024large,UserBehavior_10.1145/3708985,Recommender_AI_10.1145/3731446}.

A growing area of research focuses on datasets meant to enhance more nuanced social and collaborative intelligence. The SOTOPIA-\(\pi\) dataset supports the interactive learning of social skills through imitation and RL based on filtered social interaction data\cite{wang-etal-2024-sotopia}. Other approaches create "societies" of LLM agents to explore emergent collaborative behaviors using established benchmarks like MMLU and MATH\cite{zhang-etal-2024-exploring}. To assess an agent's ability to recognize implicit human needs, the IN3 benchmark tests proactive clarification skills, while the OEDD corpus presents scenarios where agents must apply different experiential information without being misled by distractions \cite{qian-etal-2024-tellmemore,george-etal-2024-probing}. These varied and increasingly sophisticated datasets are essential for developing agents that excel in tasks and are grounded, reliable, and socially aware.

Benchmarks and datasets play a crucial role in evaluating LLM-based agents by directly testing key capabilities such as reasoning, planning, tool use, collaboration, and generalization. For reasoning, the GSM8K dataset evaluates mathematical problem-solving. The CORY framework\cite{ma2024coevolving} improves policy optimality by reducing distribution collapse, showing strong deductive reasoning. Similarly, HumanEval and MBPP\cite{bai2025collaboration} test code generation.

In terms of planning, ALFRED\cite{song2023llm} assesses agents few-shot performance . LLM-Planner’s few-shot performance reveals grounding challenges. Similarly, TIME-ARENA\cite{zhang-etal-2024-timearena} examines time-aware multitasking in cooking and lab tasks and highlights LLMs’ difficulties with temporal dependencies.

AGENTVERSE~\cite{chen2023agentverse} evaluates text understanding and coding, showing emerging cooperative behaviors. 
SOTOPIA-\(\pi\)~\cite{wang-etal-2024-sotopia} uses MMLU to achieve social goal completion, matching GPT-4, which emphasizes social intelligence. Tool use is assessed through ToolBench~\cite{qin2024toolllm}, which tests API handling with strong zero-shot generalization. 
RE-Bench~\cite{wijk2025rebench} shows that LLMs outperform human experts in time-constrained R\&D tasks, indicating effective decision-making.

Multimodal grounding is evaluated through VQAv2 and Text-VQA. CogAgent\cite{hong2024cogagent} excels in GUI navigation, which evaluates agents' visual-language integration. By directly examining the strengths of LLM, these benchmarks also uncover limitations, such as long-term focus issues\cite{zhang2024chain} and social biases\cite{liu2025exploring}. 
Although datasets in LLM-based autonomous agent research have significantly advanced areas such as reinforcement learning, code generation, healthcare, and urban planning, they still present major limitations that hinder generalization and real-world applicability. Many existing datasets are built within simulated or synthetic environments, including ALFRED\cite{song2023llm}, Overcooked AI\cite{zhang2024proagent}, and Sociodojo\cite{cheng2024sociodojo}. These settings often fail to capture the inherent noise, unpredictability, and multimodal complexity of real environments, leading to overfitting and poor performance\cite{Understanding_the_Weakness_10.1145/3637528.3671650,zhang2024building,qiao2024agent}. Robotics and navigation datasets, for instance Street View\cite{schumann2024velma} and AndroidArena\cite{Understanding_the_Weakness_10.1145/3637528.3671650} often lack diversity in user behavior, scene variation, and environmental dynamics, resulting in biased models with limited robustness\cite{singh2024personal,yang2024llm}. Standard benchmarks, including HumanEval, MBPP, GSM8K, MATH, and WebShop, remain task-specific with limited cross-domain and multilingual coverage \cite{shinn2023reflexion,bai2025collaboration,ma2024coevolving,li2024fg}.

Existing benchmarks and datasets inadequately measure agentic reasoning by testing performance in simplified, static environments that fail to reflect real-world complexity. They often prioritize subjective, language-level evaluation over assessing an agent's objective, action-level impact on achieving goals \cite{li2025eliciting}. This reliance on metrics like final success rate offers few insights into the actual reasoning process\cite{chang2024agentboard} and neglects crucial challenges like navigating dynamic action spaces\cite{Understanding_the_Weakness_10.1145/3637528.3671650}. Furthermore, these benchmarks often overlook critical real-world constraints, such as temporal dynamics, which are essential for assessing an agent's ability to plan and multitask efficiently \cite{zhang-etal-2024-timearena}.

\vspace{-10pt}
\section{Discussion}
\label{10_discussion}
Our review explored the rapid use of LLMs as agents and tools for complex autonomous tasks. This study presents a comprehensive examination of existing LLM-based frameworks and discusses cognitive and operational components critical to agentic intelligence. By analyzing a broad spectrum of prior work, we identified how these systems are architected, the capabilities that underpin their autonomy, and the current limitations constraining their scalability. We categorized the most widely adopted methods and their implementation within single-agent and multi-agent systems across various application domains. In general, this review offers a coherent foundation for understanding the developmental trajectory of LLM agents and tools to guide future advancements in this rapidly evolving field.

\vspace{0.3em} \noindent \textbf{\textit{Baseline LLMs.}} Recent studies addressing \textit{(RQ1)} reveal that GPT-4 is currently the dominant foundational model, cited in fifty-five studies, followed by GPT-3.5 in 23 studies, establishing it as a performance benchmark despite cost and access restrictions. GPT variants fulfill three roles: first, as gold-standard ablation references; second, as multi-agent collaborators with open models; and third, as primary reasoning modules within agent stacks.

In addition to OpenAI models, Anthropic's Claude 3 series is a prominent proprietary alternative \cite{zhang2024chain, chang2024agentboard, 10802322_SMART-LLM, zhang2024privacyasst}. Google’s Gemini excels at very-long-context reasoning for document-scale planning. Meanwhile, the open-source ecosystem is rapidly advancing, reducing the performance gap. Models such as LLaMA-2/3 \cite{zhang2024building, zheng2024steveeye, wu2024chateda, xie2024can, ning2024urbankgent}, Mistral \cite{jin2024large, zhang-etal-2024-timearena, qian-etal-2024-tellmemore, wang-etal-2024-sotopia, abdaljalil-etal-2025-theorem, qiao2024agent, zhai2024fine}, Gemini \cite{zhang2024privacyasst, barra2025prompt}, Qwen, DeepSeek, Vicuna, WizardLM, and GLM-4 are increasingly adopted for coding, planning, and dialogue. Specialized derivatives like CodeLLaMA and ToolLLaMA demonstrate, task-specific pre-training can produce domain expertise without complete retraining. For constrained environments, Phi-2, Phi-3.5 Mini \cite{li2025llm}, Gemini-2B, and Mistral-7B, are more efficient. Lastly, multimodality through CLIP, Stable Diffusion XL, and VQ-GAN enables LLM agents beyond text to vision and interface automation.

\vspace{0.3em} \noindent \textbf{\textit{External tool integration in LLMs.}} We found that integrating external tools \textit{(RQ2)} is a primary driver of LLM autonomy. LLMs paired with rich computational resources, web APIs, knowledge graphs, code execution, RESTful service libraries \cite{qin2024toolllm}, and high-fidelity simulators elevate agent capability. Real-time data retrieval from domain-specific sources strengthens RAG accuracy while speech recognition tools, such as Speechly, enable real-world interaction. Furthermore, environments such as AI2-THOR, ALFWorld, and SMAC, connected through platforms like ROS 2 and Gazebo, demonstrate that LLMs can perceive state, reason, and act in dynamic and interactive settings \cite{Describe_10.5555/3666122.3667602, shinn2023reflexion, li2025llm, zhou2024star, frering2025integrating}. Collectively, these patterns strongly suggest that external tool access is an essential foundational mechanism for accessing up-to-date knowledge, enabling perception, action, and reasoning in modern LLM agents.

\vspace{0.3em} \noindent \textbf{\textit{Frameworks for LLM agents.}} Our analysis of LLM agent frameworks \textit{(RQ3)} reveals single-agent scenario is dominated by ReAct and Reflection, prioritizing streamlined reasoning-action integration and iterative self-improvement. ReAct’s thought-action sequences enable dynamic task adaptation, while Reflexion refines performance through memory-based learning and feedback \cite{shinn2023reflexion}. These frameworks excel at tasks that require adaptability, minimal infrastructure, and self-guided correction.

Multi-agent implementations favor frameworks such as AutoGen and CAMEL, effective for interaction, role differentiation, and collective planning \cite{wu2024autogen}. AutoGen’s conversation programming supports asynchronous, modular communication workflows. CAMEL’s role-conditioned framework enables agents to solve tasks collaboratively through structured, multi-turn dialogues. These capabilities are critical for multistep reasoning, distributed system design, or collaborative creativity, where agents operate semi-independently while maintaining group objectives. LangChain is compatible in single and multi-agent setups with chain reasoning steps, memory access, and tool use \cite{huang2023benchmarking, klein2025fleet}. MetaGPT \cite{hong2023metagpt} integrates software engineering conventions among less prevalent frameworks, while AIDE \cite{wijk2025rebench} and BDI \cite{frering2025integrating} focus on interpretability and logical behaviors to enhance explainability and work in rule-constrained domains.

The trend is shifting from instruction-bound architectures to adaptive, role-aware, and coordination-driven systems. The widespread use of ReAct, Reflexion, LangChain, and AutoGen indicates a convergence on reasoning, memory integration, and collaborative execution as core capacities.
\begin{figure*}[ht!]
  \centering
  \includegraphics[scale=0.1]{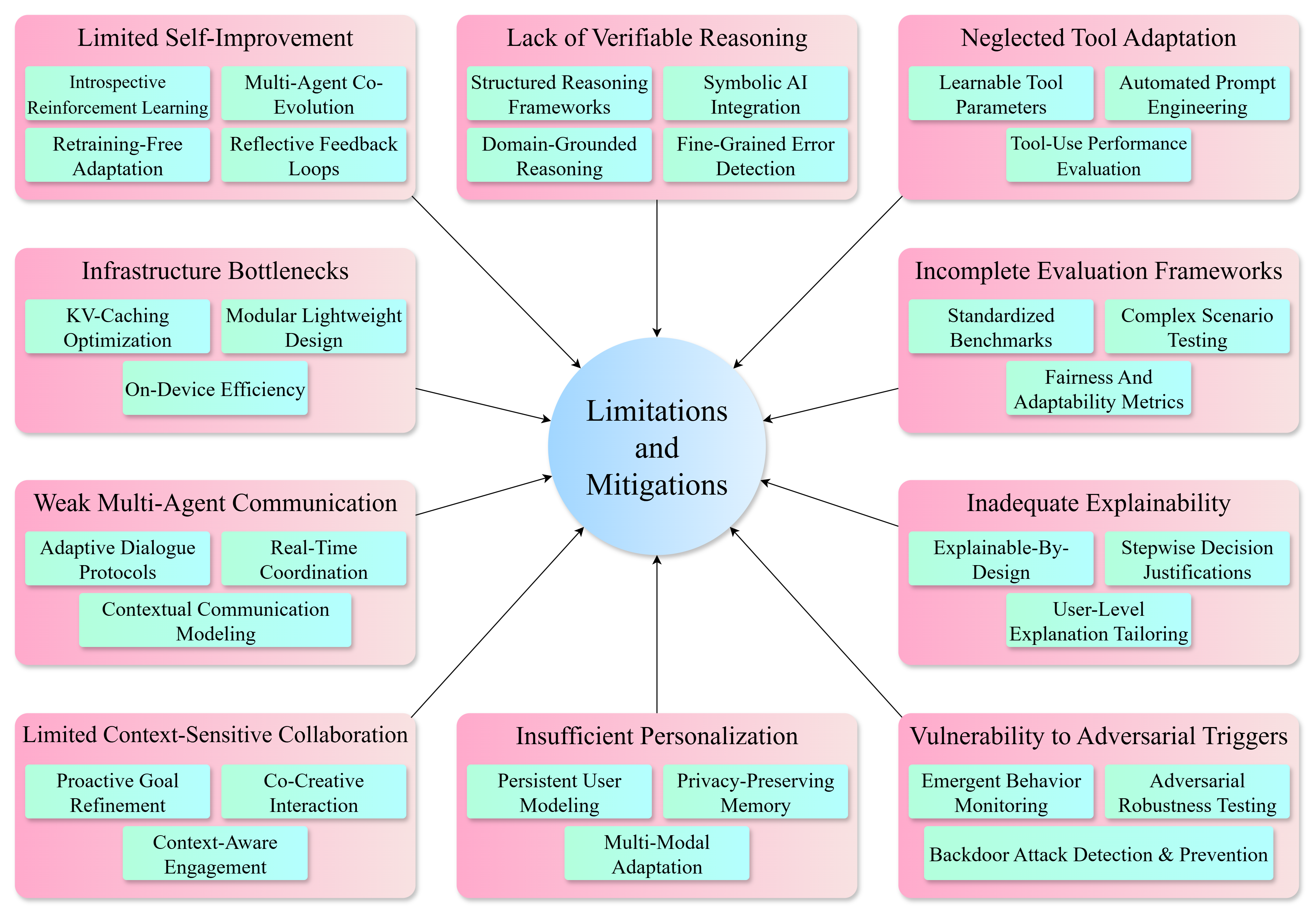}
  \caption{Limitations in the current landscape of LLM-based agents and corresponding mitigation strategies}
  \label{fig:limitations_mitigations}
\end{figure*}

\vspace{0.3em} \noindent \textbf{\textit{Reasoning, planning, and memory.}} Integration of reasoning, planning and memory mechanisms \textit{(RQ4)} shaped the development of LLM-agents. During analysis, we found single-agent frameworks like TOOLLLM \cite{qin2024toolllm}, EASYTOOL \cite{yuan-etal-2025-easytool}, and ToTh \cite{abdaljalil-etal-2025-theorem} emphasize autonomy, safety, and introspective capability. In contrast, multi-agent frameworks like AGENTVERSE \cite{chen2023agentverse}, Thread Memory \cite{jin2024large}, and ToM \cite{li-etal-2023-theoryofmind} prioritize coordination, belief modeling, and contextual reasoning. AGENTVERSE implements collaborative reasoning using structured symbolic modules; Thread Memory preserves context in multi-turn dialogue and interaction; ToM enhances coordination by modeling belief states in multi-agent settings. CoT, ReAct, and Self-Reflection appear in both. CoT is effective in code generation and robotics \cite{bai2025collaboration, li-etal-2023-theoryofmind}, where ReAct \cite{zhang-etal-2024-timearena,singh2024personal} and Reflexion \cite{zhang-etal-2024-timearena, feng-etal-2024-large} integrate feedback-aware reasoning, contributing to real-time adaptability and iterative self-improvement.

Single agent planning strategies, such as ReHAC \cite{feng-etal-2024-large} and RAFA \cite{Reason_for_future_10.5555/3692070.3693331}, emphasize proactive multistep forecasting and heuristic goal decomposition. Multi-agent systems such as DeCoAgent \cite{jin2024decoagent} and SMART-LLM \cite{10802322_SMART-LLM} address coordination through prompt-based decomposition and coalition formation, respectively. In particular, adaptive and evolution-based planning in Star-Agents \cite{zhou2024star} and UrbanKGent \cite{ning2024urbankgent} reflect increasing interest in long-horizon self-organizing workflows.

We observe a clear divide between localized and distributed memory. Single-agent models optimize within episode retention using episodic reprompting \cite{zheng2024steveeye}, compressed logs \cite{huang2023benchmarking}, or symbolic graphs \cite{abdaljalil-etal-2025-theorem}, supporting isolated task continuity. Multi-agent models like ProAgent \cite{zhang2024proagent} provide shared task memory and thread-level history, while MATRIX \cite{Selfalignment_10.5555/3692070.3693667} generates diverse interaction data as a multi-agent simulation environment, rather than managing shared memory. Context windows \cite{zhang2023large, pang2024kalm} and dialogue history \cite{wang-etal-2024-sotopia, li2023camel} underpin both categories, while hybrid memory systems \cite{boiko2023autonomous, gao2024large} signal convergence toward unified architectures that bridge short/long-term retention.

\vspace{0.3em} \noindent \textbf{\textit{Prompting, fine-tuning, and memory augmentation.}} Our findings showed that agent autonomy is enabled by prompt engineering, fine-tuning, and memory augmentation \textit{(RQ5)}. Prompt engineering, a non-parametric technique, allows flexible, inference-time control over agent behavior, as seen in  MechAgents \cite{ni2024mechagents} and AGENTVERSE \cite{chen2023agentverse}, which used structured prompts to simulate complex interactions such as expert collaboration, role play, trust modeling, and cognitive biases. However, in special domains where prompting is insufficient, fine-tuning is employed as a parametric approach to internalize domain expertise and task-specific skills, as seen in ToolLLM \cite{qin2024toolllm} for tool use specialization, RISE for self-correction abilities, and SOTOPIA-\(\pi\) \cite{wang-etal-2024-sotopia} for modeling social intelligence. Likewise, memory augmentation through external RAG and internal episodic memory is essential to ground agents in real-world contexts and support long-term reasoning. For example, TRIAGEAGENT \cite{lu2024triageagent} utilizes medical knowledge retrieval to mitigate hallucination, while Reflexion \cite{shinn2023reflexion} enables agents to adapt based on accumulated experience. Therefore, the architectural and behavioral autonomy of modern LLM agents is largely scaffolded by design choices rather than innate model capabilities.

\vspace{0.3em} \noindent \textbf{\textit{Evaluation and benchmarks.}} Our analysis indicated that evaluating LLM-based agents \textit{(RQ6)} requires a methodological change, from static NLP benchmarks to dynamic process-oriented evaluations. We also found that a robust evaluation framework should incorporate final outputs and the full agentic process. For example, AGENTBOARD \cite{chang2024agentboard} introduces "progress rate" metrics that capture incremental task completion across intermediate steps. Similarly, RE-Bench \cite{wijk2025rebench} aligns agent performance with human expert benchmarks on complex research problems, reinforcing the importance of human-level baselines in agent evaluation. 

In addition to this, controlled simulations reveal handling unreliable or adversarial collaborators \cite{huang2025ontheresilience}, highlighting resilience as a critical component of agent intelligence. Domain-specific evaluations also emerged as essential for assessing practicality. For example, ChemCrow \cite{m2024augmenting} benchmarks autonomous chemical synthesis against metrics grounded in scientific outputs, while tool-integrated agents in healthcare significantly reduce clinical calculation errors \cite{goodell2025large}.


Our analysis points to the growing recognition of failure analysis and safety evaluations as a key to agent benchmarking. Adversarial testing techniques are being developed to expose vulnerabilities, including backdoor attacks \cite{yang2024watch}, while new approaches such as PrivacyAsst \cite{zhang2024privacyasst} aim to maintain data confidentiality and trustworthiness during agent deployment. The evaluation state for LLM agents is rapidly evolving towards more comprehensive, interactive, and context-aware approaches. These emerging strategies emphasize whether an agent can succeed and how it reasons, interacts, adapts, and safeguards users along the way.

\section{Future directions}
\label{11_future_dir}
LLM-based agents demonstrate significant potential to automate complex tasks, yet their current limitations (see Figure \ref{fig:limitations_mitigations}) highlight the need for systemic research. To build agents that are reliable, adaptive, and ethically aligned, challenges such as unverifiable reasoning, limited contextual collaboration, vulnerability to adversarial attacks, and inadequate personalization must be addressed. 

In this section, we outline the key areas where future work and critical research is needed \textit{(RQ7)}. Table S3, in the Supplementary Material, provides an organized summary of the existing research gaps and potential research pathways discussed in this section.

\subsection{Towards verifiable reasoning and robust self-improvement}

Although current agents based on LLMs show impressive abilities, their reasoning processes frequently remain unclear, and their capacity to learn from failure is limited. This lack of clarity and inflexibility limits their usage. Future work should develop agents with clear, verifiable reasoning methods and strong, efficient self-improvement cycles.

One major trend in improving reasoning is shifting from unstructured, free-form "CoT" to more organized and logically sound frameworks. Some methods ensure logical consistency from the start. For example, the ToTh model's reasoning is a multi-agent collaboration that simulates abductive, deductive, and inductive inference, assembling outputs into a formal reasoning graph, validating for coherence \cite{abdaljalil-etal-2025-theorem}. In future works, a similar approach could assist in detecting financial fraud or security breaches by producing a reasoning graph that will provide a transparent, stepwise logic trail that authorities can trace and validate. MechAgents \cite{ni2024mechagents} utilizes a team of agents for complex engineering challenges and self-correcting physics-based simulation code. This verifies the reasoning of the agent against established scientific models. Building on this strategy, future work may explore the potential drug-drug interactions by modeling their combined molecular effects. Combining LLMs with traditional symbolic AI approaches, such as Belief-Desire-Intention agents, enables verifiable decision-making and improves safety \cite{frering2025integrating}. An essential aspect of verifiability is identifying and addressing flaws in the reasoning process. FG-PRM research \cite{li2024fg} reliably tackles this by creating a classification of errors in mathematical reasoning and training a reward model to identify and penalize these errors.

Alongside verifiable reasoning is the goal of strong self-improvement, where agents learn from experience without needing retraining. Through strategies like Reflexion, agents can reflect on their performance and learn from mistakes, using verbal RL, creating a memory of these reflections that guides future attempts \cite{shinn2023reflexion}. Studies such as RISE employ recursive introspection to fine-tune models through an ongoing process of self-correction, to improve their output over time \cite{xi2025llmagentsurvey}. Subsequent work could apply these strategies to develop an agent for biomedical image segmentation that self-improves iteratively, minimizes reliance on continuous large-scale human annotation, and adapts to patient-specific image differences. Multi-agent dynamics also serves as a robust method for promoting self-improvement. In the CORY framework, a model is split into pioneer and observer agents that co-evolve through a cooperative RL process. This improves resilience and avoids policy failures common in traditional RL fine-tuning \cite{ma2024coevolving}. Other frameworks, such as COPPER \cite{bo2024reflective}, feature a special reflector agent that provides tailored, constructive feedback to other agents, improving teamwork through a systematic self-improvement loop. Some studies, such as AgentOptimizer \cite{Offline_10.5555/3692070.3694566}, suggest treating agent tools as learnable parameters, enabling task adaptation through tool improvement. Likewise, AVATAR \cite{wu2024avatar} automates the optimization of prompts for tool use by employing a comparator module to reason over both positive and negative examples. Integrating structured reasoning, introspective learning, and adaptive capabilities is driving LLM agents toward more reliability, transparency, and continual improvement.

\subsection{Towards scalable, adaptive, and collaborative LLM-based agent systems}

Future research should enhance LLM agents' scalability, adaptability, and real-time operational capacity, for deployment in multimodal and domain-specific environments \cite{ColaCare_10.1145/3696410.3714877, yang2024talk2care, zeng2024gesturegpt, UserBehavior_10.1145/3708985}. Beyond internal agent reasoning and learning mechanisms, future work should address infrastructure constraints, memory usage, and latency, while enabling efficient inference through architectural techniques such as KV caching and optimized decoders \cite{jin2024large, zeng2024gesturegpt}. These improvements are vital for low-resource or time-sensitive scenarios, such as on-device processing or streaming interaction.

Another key priority lies in enhancing multi-agent collaboration and the development of perceptually aware communication. The absence of adaptive communication protocols and contextual modeling in frameworks such as AGENTVERSE \cite{chen2023agentverse} limits their ability in dynamic multi-agent environments \cite{wijk2025rebench}. Future agent systems should adopt human-inspired, socially grounded dialogue strategies to enhance collaboration capability, contextual awareness and intent inference among agents \cite{kim2024mdagents}. Such improvements will enable the deployment of lightweight agents on edge hardware for traffic systems, warehouses, and drones, facilitating inter-agent communication and collaboration.

Furthermore, as the range of agent applications expands, extending to gesture-based interfaces, EHR prediction, collaborative games, continuous learning, coevolutionary training, and prompt refinement methods become essential \cite{ColaCare_10.1145/3696410.3714877}. Simultaneously, development of negotiation-heavy simulations like Lewis-style coordination games or Diplomacy can facilitate deeper understanding of agent reasoning under critical settings \cite{wijk2025rebench, wang-etal-2024-sotopia}.

Likewise, ensuring robustness and interpretability remains a primary concern. Beyond introspective reasoning and verification, transparent evaluation pipelines are needed to track agent behavior over time, particularly under unpredictability and emergence. The acceleration of open-source advancements, led by models such as LLaMA-2-70B, provides an accessible, scalable, reproducible foundation, enabling researchers to collectively develop safe, effective, socially aligned agentic systems \cite{li2025eliciting, yin-etal-2024-agentlumos}.

\subsection{Deepening the human-agent symbiosis: personalization, proactivity, and trust}

Future research should enhance autonomy, alignment, and practical deployability of LLM-based agents, for complex, dynamic, open-ended, and high-stakes environments. Increased operational freedom produces emergent effects, from strategic cooperative gains to unsafe shortcut-seeking actions. Addressing this duality requires robust alignment strategies to mitigate adverse behaviors while preserving constructive autonomy to operate efficiently and effectively.

Simultaneously, the strategic focus is shifting toward human-centered augmentation, emphasizing collaboration, reliability, and personalization. Realizing this calls for advances in intent recognition and co-creative interaction, where agents can engage in context-sensitive conversations that anticipate user needs and refine task goals interactively, an approach illustrated by studies like "Tell Me More" \cite{qian-etal-2024-tellmemore} and ReHAC \cite{feng-etal-2024-large}.

Long-term personalization represents a critical research frontier in which future agents should build persistent, evolving user models over time while addressing memory and privacy, as stated in studies like PrivacyAsst \cite{zhang2024privacyasst}. Achieving such personalization requires integrating multimodal capabilities while aligning interface modalities to user-specific preferences. Recently, personalized LLM-based agents have introduced complex ethical and societal risks, anthropomorphism, and over-trust. Their humanlike interactions create deception and encourage vulnerable populations, such as children or the elderly, to disclose sensitive information \cite{peter2025benefits}. Since LLM-based agents lack moral agency, attributing accountability is conceptually flawed \cite{meier2025balancing}. Beyond interface risks, agent architectures suffer from jailbreaking \cite{shen2024anything}, and environmental signal-triggered backdoor attacks \cite{yang2024watch}. Jailbreaking refers to deliberate overriding of built-in safety measures of LLMs \cite{ouyang2022training}, and since agents need to handle multi-round dialogues and multiple sources of information, they become more prone to jailbreaking attacks. Effective human-agent collaboration requires both resilient safeguard frameworks against vulnerabilities and transparent, explainable reasoning \cite{frering2025integrating}.

Continuous model optimization across collaborative decoding, lightweight modular execution strategies, and efficient caching mechanisms drives real-world scalability and enables a new class of agents that go beyond technical proficiency to ethically aligned, emotionally aware, privacy-conscious, and deeply integrated in complex workflows.

\section{Conclusion}
\label{12_concl}
In this article, we present a comprehensive overview of LLM-based agents and the integration of tools within these systems. We examined how prompt engineering, fine-tuning, memory enhancement, and tool use contribute to the building of LLM agents by addressing seven focused research questions.
We observed that single-agent systems prioritize autonomy and introspective decision making. However, multi-agent systems focus on coordination, role distribution, and collaborative planning. Crucially, this distinction becomes domain sensitive, with multi-agent configurations demonstrating pronounced advantages in areas requiring social intelligence, cooperative problem solving, and high-stakes decision support, such as healthcare, scientific research, and complex engineering tasks. Integrating external tools, real-time data sources, and multimodal systems has become essential to enable LLM agents to perform tasks beyond the limitations of pre-trained models. In parallel, the evaluation of these agents is changing from static accuracy-based benchmarks to dynamic process-oriented methods that account for reasoning quality, adaptability, and task completion in real-world settings.

We have also examined critical limitations and safety concerns as LLM agents are deployed in sensitive environments. These include risks related to security, performance limitations, adaptability in dynamic or personalized settings, and challenges in trust, explainability, and agent co-evolution. Addressing these challenges will ensure reliability, trust, and efficiency in future applications.

In our view, future work should focus on two critical goals: making agent reasoning transparent and verifiable and developing reliable methods for self-improvement without compromising safety. These capabilities are critical in high-risk environments, where errors can have serious consequences. Ensuring that LLM agents are trustworthy, resilient, and aligned with domain-specific requirements will be central to their responsible deployment across disciplines.

\section*{Declarations}
\noindent
\textbf{Conflict of interests:} On behalf of all authors, the corresponding author states that there is no conflict of interest.\\
\textbf{Funding:} No external funding is available for this research.\\
\textbf{Data availability statement:} Not applicable.\\
\textbf{Ethics approval and consent to participate}. Not applicable.\\
\textbf{Informed consents:} Not applicable.\\

\textbf{Author Contributions:}
\textit{Conceptualization} Sadia Sultana Chowa, Riasad Alvi, Subhey Sadi Rahman, Mohaimenul Azam Khan Raiaan;
\textit{Methodology:} Sadia Sultana Chowa, Riasad Alvi, Subhey Sadi Rahman, 
\textit{Resources and Literature Review:} Sadia Sultana Chowa, Riasad Alvi, Subhey Sadi Rahman, 
\textit{Writing – Original Draft Preparation:} Sadia Sultana Chowa, Riasad Alvi, Subhey Sadi Rahman, Md Abdur Rahman, Mohaimenul Azam Khan Raiaan;

\textit{Validation: } Sami Azam, Mohaimenul Azam Khan Raiaan, Md Abdur Rahman, Md Rafiqul Islam;

\textit{Formal Analysis:} Sadia Sultana Chowa, Sami Azam, Mohaimenul Azam Khan Raiaan, Md Abdur Rahman;

\textit{Writing – Reviewing and Finalization:} Mohaimenul Azam Khan Raiaan, Sami Azam, Md Rafiqul Islam, Mukhtar Hussain;

\textit{Project Supervision:} Mohaimenul Azam Khan Raiaan, Sami Azam;

\end{document}